\documentclass[sigconf]{acmart}
\usepackage[symbol]{footmisc}
\usepackage{graphicx}
\usepackage{subfigure}
\usepackage{stfloats}
\usepackage{xcolor}
\usepackage{float}
\usepackage{subcaption}
\usepackage{MnSymbol}
\usepackage{threeparttable}
\hypersetup{colorlinks=true,linkcolor=blue,urlcolor=blue}
\usepackage{amsmath}

\usepackage[most]{tcolorbox}
\definecolor{deepviolet}{RGB}{96, 0, 96}

\tcbset{
  promptbox/.style={
    colback=white,
    colframe=violet,
    colbacktitle=deepviolet,
    fontupper=\itshape,
    enhanced,
    center,
    attach boxed title to top left={yshift=-0.05in,xshift=0.15in}, 
    boxed title style={boxrule=0pt,colframe=white},
    before=\vskip1.0ex, 
    after=\vskip1.0ex,  
    boxsep=2pt, 
  }
}

\newtcolorbox{promptframed}[2][]{promptbox,title=#2,#1}

\newcounter{promptcounter}

\AtBeginDocument{%
  }

\copyrightyear{2025}
\acmYear{2025}
\setcopyright{acmlicensed}\acmConference[KDD '25]{Proceedings of the 31st ACM SIGKDD Conference on Knowledge Discovery and Data Mining V.1}{August 3--7, 2025}{Toronto, ON, Canada}
\acmBooktitle{Proceedings of the 31st ACM SIGKDD Conference on Knowledge Discovery and Data Mining V.1 (KDD '25), August 3--7, 2025, Toronto, ON, Canada}
\acmDOI{10.1145/3690624.3709231}
\acmISBN{979-8-4007-1245-6/25/08}

\begin{document}

\title{CausalMob: Causal Human Mobility Prediction with LLMs-derived Human Intentions toward Public Events}

\author{Xiaojie Yang}
\affiliation{%
  \institution{The University of Tokyo}
  \city{Tokyo}
  \country{Japan}
}
\email{xiaojieyang@g.ecc.u-tokyo.ac.jp}

\author{Hangli Ge}
\affiliation{%
  \institution{The University of Tokyo}
  \city{Tokyo}
  \country{Japan}
}
\email{hangli.ge@koshizuka-lab.org}

\author{Jiawei Wang}
\affiliation{%
  \institution{The University of Tokyo}
  \city{Tokyo}
  \country{Japan}
}
\email{jiawei@g.ecc.u-tokyo.ac.jp}

\author{Zipei Fan}\authornote{Corresponding author. \dag \! First affiliation.}
\affiliation{%
  \institution{School of Artificial Intelligence\\
  Jilin University\dag}
  \city{Changchun}
  \country{China}
  }
\email{fanzipei@jlu.edu.cn}

\author{Renhe Jiang}
\affiliation{%
  \institution{The University of Tokyo}
  \city{Tokyo}
  \country{Japan}
}
\email{jiangrh@csis.u-tokyo.ac.jp}

\author{Ryosuke Shibasaki}
\affiliation{%
  \institution{The University of Tokyo}
  \city{Tokyo}
  \country{Japan}
}
\email{shiba@csis.u-tokyo.ac.jp}

\author{Noboru Koshizuka}
\affiliation{%
  \institution{The University of Tokyo}
  \city{Tokyo}
  \country{Japan}
}
\email{noboru@koshizuka-lab.org}

\renewcommand{\shortauthors}{Xiaojie Yang et al.}

\begin{abstract}
Large-scale human mobility exhibits spatial and temporal patterns that can assist policymakers in decision making. Although traditional prediction models attempt to capture these patterns, they often interfered by non-periodic public events, such as disasters and occasional celebrations. Since regular human mobility patterns are affected by these events, estimating their causal effects is critical to accurate mobility predictions. News articles provide unique perspectives on these events, though processing is a challenge. In this study, we propose a causality based prediction model, \textbf{CausalMob}, to analyze the causal effects of public events. We first utilize large language models (LLMs) to extract human intentions from news and transform them into features that act as causal treatments. Next, the model learns representations of spatio-temporal regional covariates from multiple data sources to serve as confounders for causal inference. Finally, we present a causal effect estimation framework to ensure that event features remain independent of confounders during prediction. Based on large-scale real-world data, the experimental results show that the proposed model excels in human mobility prediction, outperforming state-of-the-art models.
\end{abstract}



\maketitle
\section{Introduction}
\begin{figure}[htbp]
\begin{center}
\includegraphics[scale=0.3]{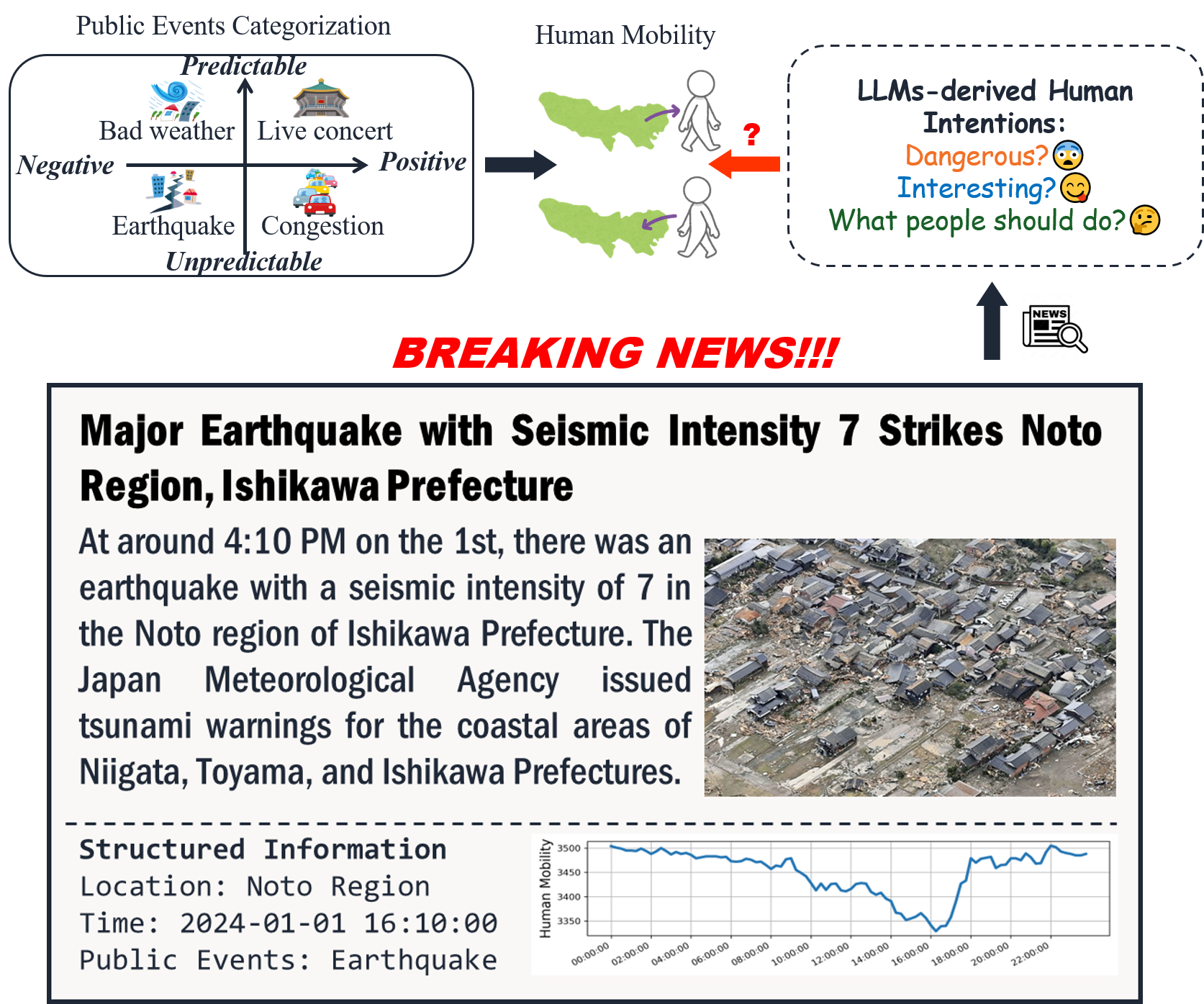}
\end{center}
\caption{Conceptual graph of this research}
\label{concept}
\end{figure}

\begin{figure}[htbp]
\begin{center}
\includegraphics[scale=0.25]{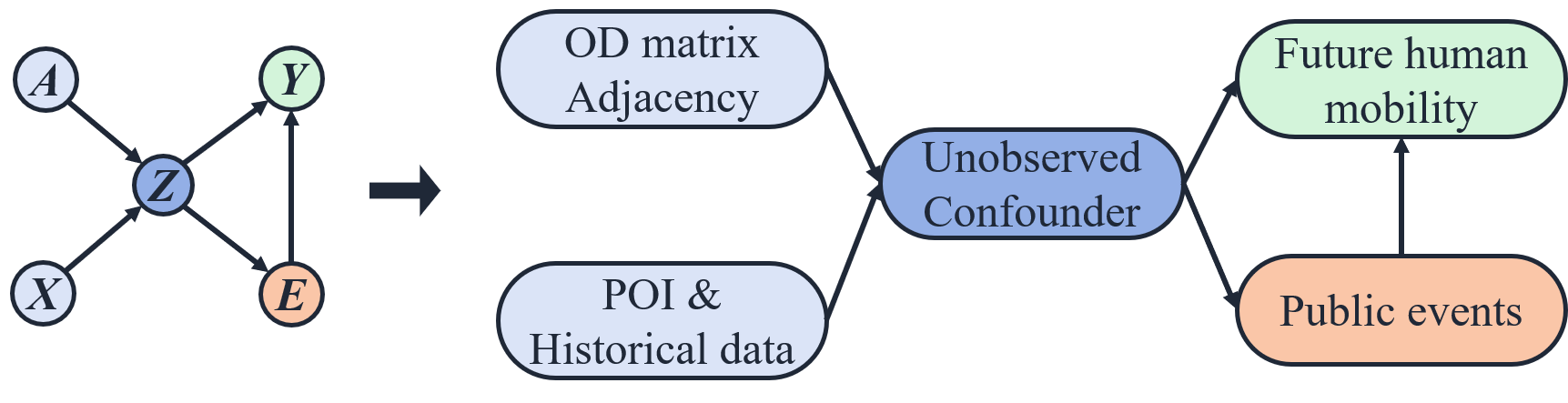}
\end{center}
\caption{Causal graph of this research}
\label{causal}
\end{figure}

Human mobility exhibits spatio-temporal patterns that can aid policy-makers in urban management and decision-making. However, large-scale public events complicate the capture of these mobility patterns, resulting in less robust predictions \cite{song2010limits,hasan2013spatiotemporal, jiang2023learning}. These public events change human mobility in various ways, driven by corresponding human intentions \cite{chen2020dualsin, jiang2022will}. For example, as shown in the "Public Event Categorization" of Figure \ref{concept}, typhoons can be predicted through weather reports, causing people to stay home. In contrast, emergencies like earthquakes have similar effects but are harder to forecast. Additionally, live concerts and traffic congestion can increase human volumes in specific regions, with the latter often links to unexpected traffic accidents.
\indent Based on the above cases, our research categorizes public events based on their predictability and their impact on human mobility (positive or negative to count increases), as shown in Figure \ref{concept} \cite{wang2015regularity}. Understanding and analyzing human mobility and intentions toward public events using this categorization criteria can help decision-makers better prepare for potential opportunities and risks. However, current research often captures the statistical dependencies between human mobility and public events without analyzing the \textit{causal relationship} between them \cite{zhiwen2023assessing}. This limitation is significant because human intentions play a crucial role in how people respond to public events. To support our point, we provide an example of an earthquake occurring on January 1, 2024, along with the corresponding human mobility in Figure \ref{concept}. Therefore, to address this limitation, we leverage LLMs to simulate human intentions toward public events, enabling an explainable exploration of the causal relationship in human mobility \cite{li2023metaagents,wang2023would}.\\
\indent However, traditional causal inference methods that use randomized controlled trials to study causal effects \cite{farmer2018application} face difficulties in abstracting public event features and controlling their occurrence. Instead, researchers overcome this challenge by leveraging counterfactual analysis to study causal effects in different scenarios \cite{morgan2015counterfactuals}. Given this background, we present a causal graph to discuss the causal relationships among variables in this research, as illustrated in Figure \ref{causal}. We aim to analyze the causal effect of public events $E$ (treatment) on human mobility $Y$ (outcomes) by distinguishing the inference of spatio-temporal regional confounders (referred to as \underline{\textit{\textbf{confounders}}} in the following content) $Z$ that we learn from observation data, including spatio-temporal covariates $X$ and Origin-Destination (OD) matrix adjacency $A$ \cite{ma2022assessing,zhiwen2023assessing}.
\indent Benefiting from the developed media industry, we can obtain high-quality reports on public events through news articles and extract structured information from them \cite{li2024causalstock, wang2024news}. This is illustrated in Figure \ref{concept}, which includes translated news articles and corresponding human mobility data that fluctuated significantly due to an earthquake. Furthermore, based on previous research, with the help of LLMs, we can extract structured information more efficiently from articles and simulate human intentions toward public events as features through prompt engineering \cite{runfeng2023lkpnr, hasan2024llm}.\\
\indent Accurate estimations of causal effects rely on the unconfoundedness assumption, which requires learning confounders with all features independent of treatments \cite{wang2022generalization}. As a weak assumption, our research employs a causal effect calculation framework to remove dependencies between public events and confounders. We begin by extracting structured information from news articles and simulating mobility-related human intentions based on prompted LLMs. Next, we learn the representations of the confounders from observational data. In the causal inference part, an inference network predicts future human mobility, and a re-weighting network helps to remove event bias for each region \cite{wang2022generalization}. Finally, we estimate the causal effect of public events on human mobility. We evaluated our proposed method on real-world data, including human mobility data and news articles, covering most major administrative regions in Japan. Our contributions are as follows:
\begin{itemize}
      \item We introduce LLMs to process news articles, extracting structured information on public events and simulating human intentions as treatment features in causal inference.
      \item We develop a causal framework to learn confounder representations and remove their influence from the effects of public events represented with multidimensional features.
      \item We conducted various experiments to demonstrate the superior performance of our proposed method and provided interpretable results with thorough analysis.
\end{itemize}
\section{Related Work}
\textbf{Human Mobility Prediction.} In recent years, the development of geolocated datasets and data mining technologies has made human mobility increasingly important for urban management, transportation and disaster response \cite{song2016prediction, hangli2022multi}. Benefiting from high-performance computational power, large-scale human mobility prediction based on deep learning has achieved good performance \cite{zhao2016urban}. Researchers have focused on capturing periodical and semantic information to represent regional features for prediction. For example, STGCN \cite{yu2017spatio} captured long-term periodical spatio-temporal patterns based on graph structures for traffic prediction, while DeepSTN+ \cite{lin2019deepstn+} modeled long-range spatial dependence of different regions using POI features as semantic information. These models effectively capture regular patterns in human mobility. However, for irregular situations with low periodicity, such as disasters or festivals, previous research tends to model mobility under specific conditions, making predictions less generalizable \cite{fan2018online, jiang2019deepurbanevent, wang2022event}.\\
\textbf{Causal Analysis for Human Mobility.} Causal analysis aims to reveal cause-effect relationships among variables \cite{pearl2009causal}. Many previous studies have separated treatments from traditional inputs to analyze their effects on outcomes. These treatments are usually single variables, including discrete and continuous variables \cite{yao2021survey}. However, in the real world, treatment variables affecting human mobility are much more complex. Previously, causal analysis for human mobility has focused on specific situations. For example, Ma \cite{ma2022assessing} introduced causal inference to assess the effectiveness of COVID-19-related policies by learning confounders from human mobility data and residents' vigilance from Google Trends. Similarly, Zhang \cite{zhiwen2023assessing} used these data to assess the effect of typhoons, represented by wind speed and precipitation, on human mobility.
\section{Data and Analysis}
In this section, we describe the observation data used to estimate the causal effects of public events on human mobility. We also include data analysis of public events and human mobility patterns to demonstrate the necessity of introducing causality analysis.
\subsection{Observational Data}
This research aims to causally analyze the relationship between public events (treatment) and human mobility (outcomes) in Japan. We select 490 administrative regions (referred to as \underline{\textbf{\textit{regions}}} in the following content) across Japan, as shown in Figure \ref{map}. Some areas were exclude due to insufficient human mobility data (aggregated from GPS records provided by BlogWatcher Inc. \cite{blogwatcher}). We continuously filter these data within the time range of 2023/04/01 to 2024/03/31, corresponding to the official fiscal year in Japan.\\
\textbf{Treatment} \textit{E}: Public events features extracted from news articles. We collaborate with Kyodo News, one of the largest international news agencies, which distributes news to almost all newspapers, radio, and television networks in Japan, reaching over 50 million subscribers \cite{enwiki:1187452593}. With the same time range as the human mobility data, we filter out unrelated articles as data-cleaning process. We then align the news articles with the corresponding regions using manually labeled region codes. To further analyze the effects on human mobility, we introduce LLMs to extract structured information and human intentions as features from these filtered news articles. More details will be provided in the methodologies section.\\
\textbf{Outcome} \textit{Y}: Future human mobility in selected regions. This study focuses on human mobility affected by the causal effects of public events. We aggregate human mobility data (GPS records) in selected regions as our prediction target. During this process, all users remain anonymous, and we take the hourly samples of \textit{uid}s aggregation within mentioned time ranges.\\
\textbf{Covariates} \textit{X}: Spatio-temporal regional covariates. We introduce proxy variables as region covariates to capture the unobserved confounders $Z$. We collect Points of Interest (POI) from OpenStreetMap (OSM) and categorize them to represent region features \cite{lin2019deepstn+}. Before calculating confounders $Z$, we concatenate POI features, time embeddings, region embeddings, and historical human mobility data. More details will be provided in the methodologies section.\\
\textbf{Networks} \textit{A}: Re-normalized Mobility network. We use a human mobility network to model the inter-regional interactions and capture unobserved confounders in the selected regions. We aggregate the hourly Origin-Destination (OD) matrix adjacency within the regional network to represent the strength of each region's relationship. Each edge in this network contains a time-varying weight computed based on mobility volumes between two regions in the OD matrix. We re-normalize this mobility network, resulting in a weighted adjacency matrix $A_{t}$ for each timestamp \cite{kipf2016semi}.
\subsection{Preliminary Data Analysis}
\begin{figure}[h!]
	\centering
	\subfigure[Research Area (Japan)]{
		\begin{minipage}[b]{0.3\textwidth}
			\includegraphics[width=1\textwidth]{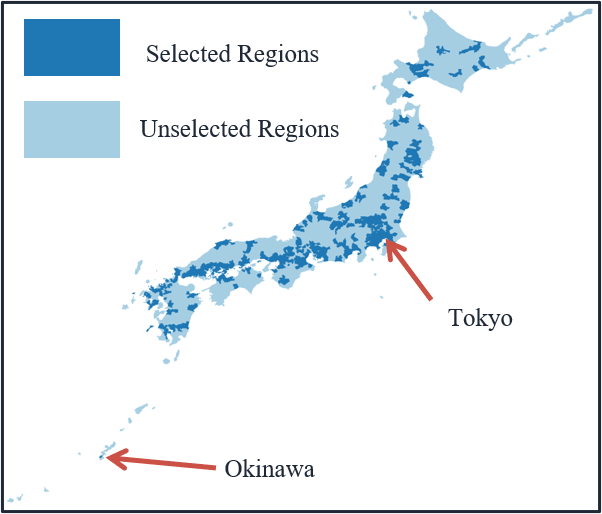}
                \label{map}
		\end{minipage}
	}\\
    \subfigure[Sumidagawa Firework (Tokyo, Sumida)]{
    	\begin{minipage}[b]{0.5\textwidth}
   		\includegraphics[width=1\textwidth]{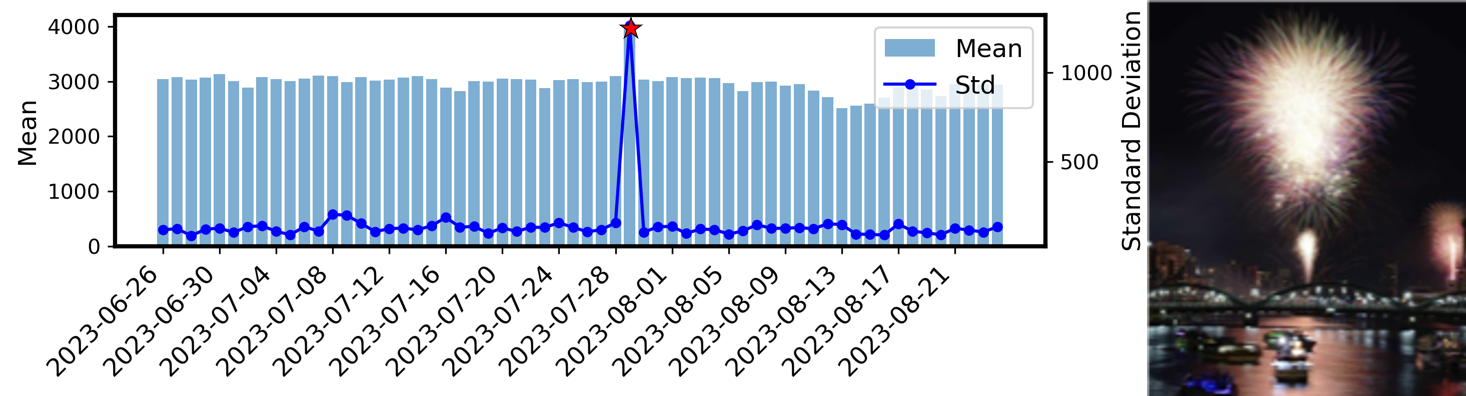}
            \label{sumidagawa}
    	\end{minipage}
    }\\
    \subfigure[Typhoon Khanun (Okinawa, Naha)]{
    	\begin{minipage}[b]{0.5\textwidth}
   		\includegraphics[width=1\textwidth]{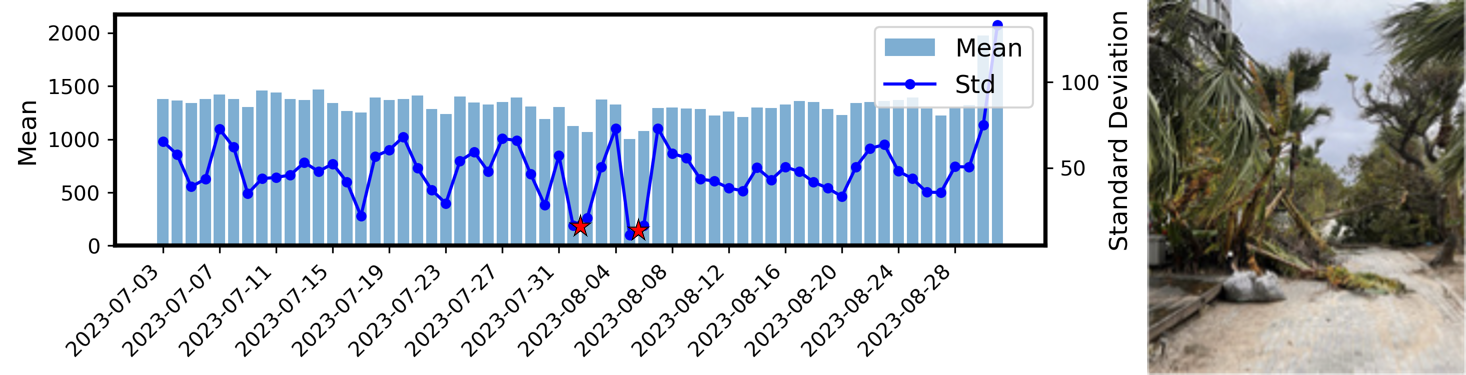}
    	\label{typhoon}
            \end{minipage}
    }
	\caption{Public events examples along with changes of daily human mobility. We calculate mean values and standard deviations of daily human mobility records ({\color{red} \large{$\star$}} markers refer to the dates when public events happened).}
	\label{cases}
\end{figure}
\noindent To explore the mentioned causal relationship, we conduct preliminary data visualization analysis to illustrate the dependence between changes in human mobility and the occurrence of public events. We manually select two cases including a firework and a typhoon from our datasets and visualize them, respectively.\\
\indent As illustrated in Figure \ref{concept}, public events are divided into four types based on their positive or negative effects on human mobility and predictability. While most cases are simple and straightforward, \textit{positive unpredictable} events are much harder to capture. We assume some might be accompanied by \textit{negative unpredictable} events or side effects of policies, such as the behavior of moving to low-cost areas due to remote working policies during COVID-19. Public events of this type require more complex reasoning processes, so we do not focus on them in this research. Furthermore, news articles are sometimes not temporally synchronized with public events, which is one of the main reasons we introduce LLMs to correct the exact occurrence time of public events, as part of structured information. Here, we provide cases as follows and related regions are marked in Figure \ref{map} with daily human mobility patterns:\\
\textbf{Sumidagawa Firework Festival}. Held at 19:00 on June 29, 2023, the Sumidagawa Firework Festival was the largest fireworks festival in Japan, attracting over one million people including locals and visitors to the related regions in 2023. As shown in Figure \ref{sumidagawa}, human mobility maintained regular patterns except on the 29th.\\
\textbf{Typhoon Khanun}. Hitting Okinawa on the 2nd and 4th of August 2023, the typhoon caused widespread power outages after its landfall. As shown in Figure \ref{typhoon}, human mobility changed significantly during the typhoon and recovered one day later.\\
\indent Based on the above cases, different public events have varying effects on human mobility, with both positive and negative impacts as we mentioned. In this research, we aim to capture these effects from a causal perspective. In this way, we can predict how human mobility will change if similar public events occur in other regions.
\subsection{Problem Definition}
In this research, we investigate the causal effects of public events on human mobility. Our methodology leverages large language models (LLMs) with chain-of-thought prompting to extract structured information, including human intentions, denoted as $N = [n_{1}, n_{2, ...}]$, from news articles. Additionally, we process aggregated human mobility data, represented as $\textbf{X}$, and mobility networks, denoted as $A$, to predict future human mobility patterns, $\textbf{Y}$, in selected regions over fixed time steps. In particular, we focus on crowd flow, measured as human volume in each region, treating it as the main mobility feature for analysis.

\section{Methodology}
In this section, we explain why and how to obtain human intentions towards public events extracted from news articles as treatments with LLMs and describe how we formulate the causal effect estimation problem in human mobility prediction.
\subsection{Human Intentions Extraction with Semantic-Aware LLMs}
\begin{figure}[h]
\begin{center}
\includegraphics[scale=0.45]{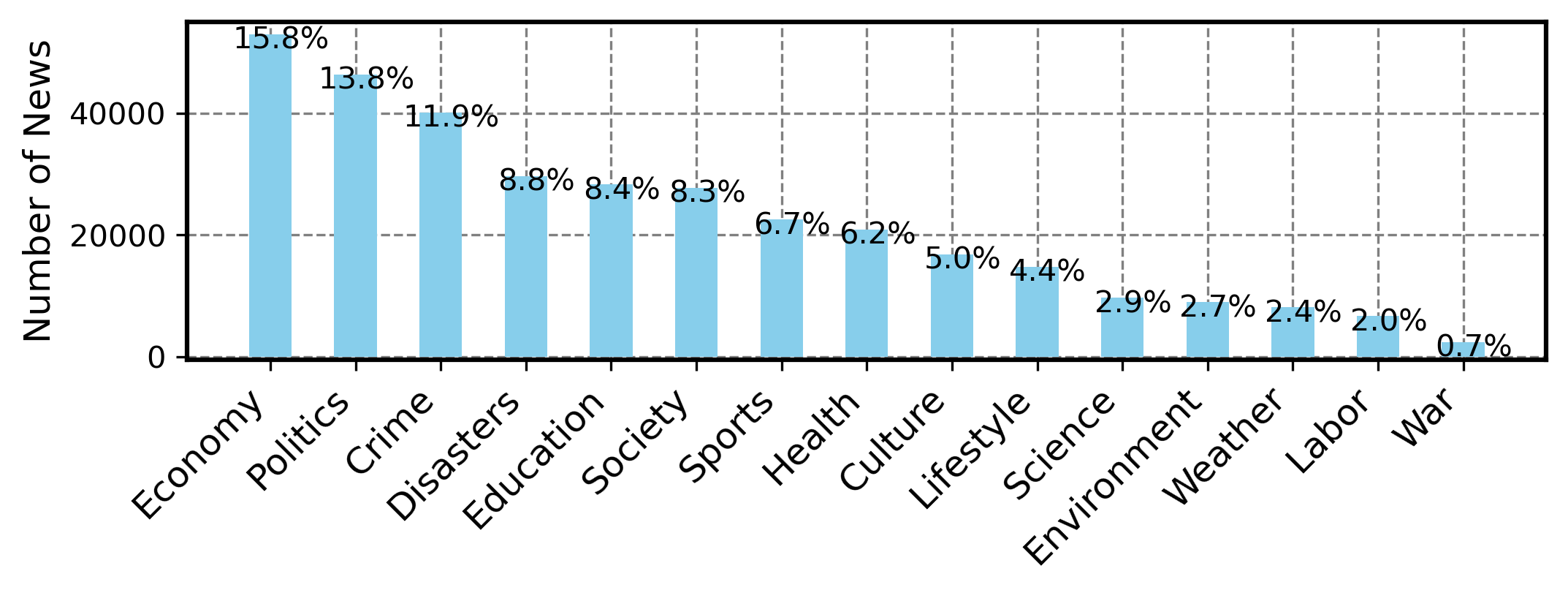}
\end{center}
\caption{Number of News labeled with categories}
\label{newscategory}
\end{figure}

In most situations, news articles report public events objectively, allowing us to estimate mobility-related human intentions. Our dataset contains over 130,000 news articles of one year, covering multiple genres as shown in Figure \ref{newscategory}, making manual processing impractical. Taking advantage of the development of open source LLMs, we can extract the necessary structured information \cite{jiang2023structgpt} and simulate human intentions related to human mobility using prompt engineering, starting with simple manual filtering \cite{jin2024comprehensive}.\\
\indent To enable LLMs to reason about human mobility-related intentions, we design a step-by-step dialog framework as shown in Figure \ref{causalmob} inspired by chain of thought \cite{wei2022chain,vijayan2023prompt}. With carefully designed prompts, we aim for LLMs to extract necessary information from news articles on the following aspects: 1) Most influential events, 2) Time of the event, 3) Four basic questions related to human mobility, i.e., where, when, who, and how (3W1H), 4) Predictability of events, and 5) Human intentions of the public events with scores. Due to space limitations, prompt details can be found in the Appendix \ref{prompt}. Historical answers will inform LLMs' future reasoning. To ensure the robustness of the generated answers for each prompt, we use one of the most powerful open-source LLMs, \textit{Llama3-70b} \cite{touvron2023llama} based on a library called \textit{Ollama} \cite{repository}. Finally, we obtain structured information of public events as shown in the top-right of Figure \ref{causalmob}. Specifically, human intentions toward public events will be used as causal treatments on human mobility in our prediction model. Furthermore, news cases are listed in Appendix \ref{newscases}.
\subsection{Causal Effect Inference through Counterfactual Estimation}\label{formulation}
\begin{figure*}[ht]
\begin{center}
\includegraphics[scale=0.6]{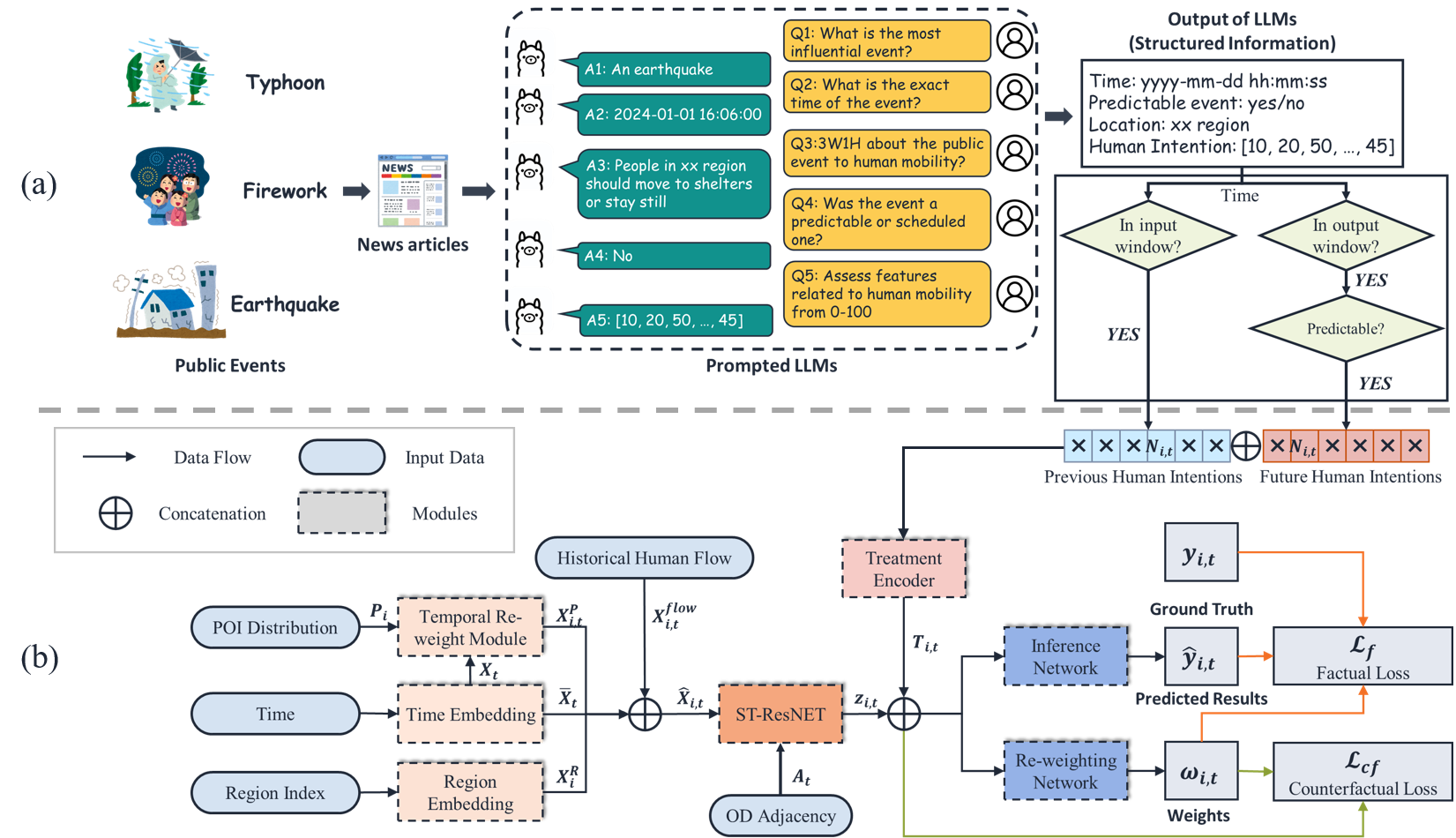}
\end{center}
\caption{Framework of CausalMob: We first process news articles to obtain structured information of public events in part (a). Then in part (b) we use ST-ResNET to learn the representation of the confounder $z$. Finally, an inference network and a re-weighting network are built to calculate factual loss and counterfactual loss as causal inference.}
\label{causalmob}
\end{figure*}
For features of $n$ selected regions, we collect POI information and divide them into $c$ specific categories according to their labels. We denote this feature as $P_{i} = \{p^{j}_{i}\}^{c}_{j=1}$, where $p^{j}_{i}$ refers to the $j$-th category of the $i$-th region. Using human mobility data consisting of GPS coordinate records, we aggregate anonymous \textit{uid}s in one region with a fixed time interval. This can be represented as $X^{flow}_{t} = \{x^{flow}_{i, t}\}^{n}_{i=1}$. Furthermore, considering that human mobility is highly dependent on regional interactions, we used a $n \times n$ normalization adjacency matrix $A_{t}$ at timestamp $t$ to represent time-varying relations between regions. In Figure \ref{causalmob}, we calculate confounder $Z_{t} = \{ z_{i, t} \}^{n}_{i=1}$ with observation data, where $X^{Reg}_{i}$ refers to natural region representations, $X_{t}$ is the embedded timestamp, and $X^{n}_{i,t}$ is temporally re-weighted POI feature.\\
\indent As for treatments for causal inference, we obtain structured information including information about time, location and human intentions toward public events of each region $N_{i, t}$ at time step $t$ with prompted LLMs, and we conclude observed events in the input window and scheduled or predictable ones in the output window, as the flow chart in Figure \ref{causalmob}. When it comes to evaluating treatment effects, one of the challenges is that previous studies focused on binary or continuous covariates in one dimension, while in our research we are facing complicated event features depicted by news articles over multiple time steps, and we need to convert them to a single feature $T_{i, t}$. To calculate the causal effect of different public events on human mobility, we need to calculate the average treatment effect (ATE) at timestamp $t$ of $i$-th selected region in our research area:
\begin{equation}
{\tau}^{i, t}_{ATE} = \mathbb{E}\left[ y^{1}_{i, t} - y^{0}_{i, t}| z_{i, t}, T_{i, t} \right],
\end{equation}
where $y^{1}_{i, t}$ and $y^{0}_{i, t}$ refer to potential outcomes of human mobility with and without causal effects of public events under the same spatio-temporal condition, corresponding to the outcomes \textit{Y}.\\
\indent Generally, it is impossible to observe the effects of different public events in one region simultaneously. Furthermore, in our observation data, public events show biases because of confounder imbalance and non-randomized treatments among regions. It means that some events tend to happen in certain regions, which will lead to unreliable inference. Processing public event features into a one-dimensional feature will result in significant information loss. Fortunately, we can quantify the \textit{similarity} or \textit{strength} of public events at a semantic level. Therefore, we introduce the concept of \textit{average dose-response function} (ADRF) \cite{imbens2000role}, denoted as $\mathbb{E}(Y_{t} | Z_{t}, T_{t} = \mathcal{T})$, to apply causal effect evaluation with multidimensional treatments. Previous research stressed that ADRF loss is bounded by the marginal loss comprising factual loss and counterfactual loss. As we cannot easily observe and define a counterfactual, we alternatively use a re-weighting schema to reduce confounder imbalance and calculate the upper bound of ADRF loss with Integral Probability Metrics (IPM) distance as our counterfactual loss (also an auxiliary loss) \cite{wang2022generalization}. As IPM distance represents a group of metrics to measure distribution differences between re-weighted fact and counterfactual, optimizing the upper bound allows us to encompass most ADRF estimation methods used in previous research.\\
\indent Based on the above formulations, we propose a causal effects calculation framework, as shown in part (b) of Figure \ref{causalmob}, to evaluate the causal effect of public events on human mobility using LLMs-derived human intentions. In our framework, we process news articles with LLMs to extract structured information, including human intentions. We then calculate spatio-temporal regional confounders using observation data within a certain input window. After aggregating the characteristics of the public event at the predicted timestamp $t$ as $T_{i, t}$ for each region, we develop a causal inference module to calculate the reweighted factual loss and the upper bound of the counterfactual loss using the IPM distance. Finally, we output regional human mobility within an output window. More details will be provided in the following section.
\subsection{Confounder Representation Learning by Spatio-Temporal Data Modeling}
In Figure \ref{causalmob}, to learn spatio-temporal regional confounders, we use ST-ResNET \cite{zhang2017deep} based on Graph Convolution Networks (GCNs) to capture hidden states independent of treatments from historical observation data. First, we capture time-varying regional covariates using POI features and time information \cite{lin2019deepstn+}. For each region, we collect POI data with $c$ categories, represented as $P_{i} \in \mathbb{R}^{n \times c}$. After normalization, we fuse POI features with embedded time information $X_{t} \in \textbf{W}^{T \times c}$ within a fixed input window. Finally, we re-weight the static POI features with average time embedding and calculate the mean value as time-varying regional features:
\begin{equation}
\textbf{X}^{P}_{i, t} = \overline{\textbf{X}}_{t} \odot \textbf{P}_{i},
\end{equation}
\noindent where $\odot$ refers to element wise multiplication and $\overline{ \textbf{X}}_{t}$ represents mean value of temporal embedding of input window.\\
\indent Next, we concatenate multiple time-step ground truth observations $\textbf{X}^{flow}_{i, t}$ with other input variables before being passed into the GCN-based ST-ResNET, just as described below \cite{he2016deep,kipf2016semi}:
\begin{equation}
\hat{\textbf{X}}_{i, t} = \textbf{X}^{flow}_{i, t} \oplus 
\textbf{X}^{P}_{i, t} \oplus  \overline{ \textbf{X}}_{t} \oplus  \textbf{X}^{Reg}_{i},
\end{equation}
\noindent where $\oplus$ refers to concatenation operations. As for $f_{GCN}$, we use a two-layer GCN to fuse and capture the hidden state as follows:
\begin{equation}
f_{GCN}(\textbf{X}, A) = A\left[(A \textbf{X})\textbf{W}_{1}\right]\textbf{W}_{2},
\end{equation}
where $\textbf{W}_{1}$ and $\textbf{W}_{2}$ are learnable parameters of two GCN layers. Finally, the output of $f_{GCN}$ will be added with input to build up a ST-ResNET \cite{zhang2017deep} structure to get the confounder $\textbf{z}_{i, t}$:
\begin{equation}
\textbf{z}_{i, t} = f_{GCN}\left(\hat{\textbf{X}}_{i, t}, A_{t}\right) + \hat{\textbf{X}}_{i, t}.
\end{equation}
\subsection{Human Mobility Prediction Enhanced by Causality Modeling}
With the confounder $\textbf{z}_{i, t}$, the model starts calculating the factual loss and counterfactual loss. We encode the treatment $T_{i, t}$ with human intentions $N_{i, t}$ in structured information. Considering that the impact of events diminishes over time, we chose GRU to process these features and use the output $T_{i, t}$ as representations of treatments. As mentioned in Section \ref{formulation}, we need to calculate re-weighted factual loss and use the upper bound of IPM distance as counterfactual loss. We use two multilayer perceptron (MLP) layers, denoted as the inference network $f_{inf}$ and the re-weighting network $f_{rwt}$, to realize the causal inference process:
\begin{equation}
w_{i,t} = {f_{rwt}}\left(\hat{\textbf{z}}_{i,t}, T_{i,t}\right),
\end{equation}
\begin{equation}
\hat{y}_{i,t}= f_{inf}(\hat{\textbf{z}}_{i,t}, T_{i,t}),
\end{equation}
where $w_{i}$ is the weight to remove regional selection bias of treatments, and $\hat{y_{i}}$ refers to predicted outcomes of future human mobility. Then, the factual loss can be represented as:
\begin{equation}
\mathcal{L}_{f} = \frac{1}{n}\sum^{n}_{i=1} \left[\left(w_{i,t} \cdot \left(\hat{y}_{i,t} - y_{i,t}\right)^{2}\right)\right].
\end{equation}
\indent On the other hand, as we aim to calculate IPM distance, we need to adjust the representations of treatments and confounders to ensure their original distributions and conditional distributions, denoted as $\textbf{z}$ and $\textbf{z}^{T}$, are independent of each other \cite{schwab2020learning}. To calculate reasonable IPM distance, we create a manually created baseline (i.e., all zero scores for human intentions) and input it into the same GRU modules to get a baseline treatment $\mathcal{T}_{b}$, simulating regular patterns without any events. We then calculate cosine similarities between the baseline treatment and all other treatments, sorting all treatments according to the results. We use $p^{j}(z)$ and $p^{j}(z|T)$ to represent original distributions and conditional distributions for the $j$-th similarity interval after dividing $[-1, 1]$ into $J$ intervals. Therefore, we can calculate the upper bound of IPM distance with maximum inter-interval as counterfactual loss:
\begin{equation}
\mathcal{L}_{cf} = \frac{1}{n}\sum^{J}_{i=1}max_{j\in\{1, 2, \dots, J\}, j \neq i}\left\{IPM_{\mathcal{G}}(p^{j}(z), p^{i}(z|T))\right\},
\label{cfloss}
\end{equation}
where $IPM_{\mathcal{G}}$ refers to the Maximum Mean Discrepancy (MMD) metric \cite{gretton2012kernel}, which is used to measure the difference between two representation distributions. Based on the above content, our final prediction target and loss function for model training is:
\begin{equation}\label{losses}
\mathcal{L} = \mathcal{L}_{f} + \alpha\mathcal{L}_{cf},
\end{equation}
where $\alpha$ controls the weight of auxiliary counterfactual loss $\mathcal{L}_{cf}$.
\section{Experiment}
In this section, we conduct experiments to introduce different analyses of our model based on Figure \ref{causalmob}. In Section \ref{performance}, we show overall performance by comparing with baseline models and present the results of the ablation study. In Section \ref{humanintentions}, we provide distributions of LLMs' output for human intentions toward public events in news articles. In Sections \ref{caseeffect} and \ref{caseevents}, we provide analyses of average treatment effects estimation and the causal response of human mobility to two public events as counterfactuals. As a case study in Section \ref{casestudy}, we show prediction results in our test process comparing predictions without human intentions.
\begin{table*}[ht]
\centering
\caption{Prediction Performance Comparison of Human Mobility Prediction Models with multiple experiments (Mean/Standard deviation). The best results are bold and the best baselines are marked with \underline{underline}, where we calculate the improvements. Ablation studies are marked with $\dag$. The time interval for each term is \textit{one hour}.}
\begin{tabular}{c|ccc|ccc|ccc}
\hline
                      & \multicolumn{3}{c|}{Short Term (In 6, Out 1)}             & \multicolumn{3}{c|}{Medium Term (In 12, Out 6)}             & \multicolumn{3}{c}{Long Term (In 24, Out 24)}              \\ \hline
Models                & RMSE         & MAE         & MAPE(\%)        & RMSE         & MAE         & MAPE(\%)        & RMSE         & MAE         & MAPE(\%)        \\ \hline
3DGCN                 & 91.50/0.47 & 38.57/0.39 & 1.72/0.03 & 275.79/5.47 & 119.97/2.56 & 5.81/0.07 & 401.63/11.41 & 175.76/2.72 & 7.98/0.20 \\
GTS                   & 88.10/10.84 & 39.41/3.61 & 2.02/0.18 & 344.68/29.99 & 181.20/6.52 & 8.87/0.12 & 427.31/29.27 & 208.65/4.77 & 9.60/0.06 \\
PromptST              & 64.29/1.25 & 26.72/0.39 & 1.65/0.02 & 105.87/2.20 & 47.51/0.51 & 2.60/0.02 & 233.92/21.12 & 81.98/4.60 & 3.98/ 0.17 \\
STGNN                 & 67.43/0.67 & 31.81/0.21 & 1.87/0.02 & 299.30/5.54 & 156.54/0.97 & 8.04/0.04 & 383.93/6.68 & 184.28/3.59 & 8.95/0.15 \\
DeepST                & 55.11/1.25 & 28.91/0.56 & 1.53/0.03 & 101.75/1.80 & 54.88/0.79 & 2.80/0.05 & 163.80/1.47 & 81.44/1.01 & 3.95/0.06 \\
DeepSTN+              & 39.45/0.45 & 22.25/0.19 & 1.21/0.01 & 87.57/2.26 & 48.28/1.24 & 2.51/0.07 & 145.92/1.32 & 70.07/1.15 & 3.43/0.05 \\
STID                  & \underline{37.19/0.96} & \underline{20.76/0.50} & \underline{1.10/0.02} & \underline{73.15/1.63} & 38.42/0.62 & 1.93/0.02 & \underline{133.80/3.15} & \underline{54.76/1.73} & \underline{2.76/ 0.12}\\ \hline
CIDER (Na\"ive)       & 56.95/2.66 & 24.92/0.68 & 1.42/0.03 & 79.89/1.94 & 36.99/0.43 & 2.06/0.02 & 139.40/1.72 & 54.98/0.51 & 2.85/0.03 \\
CIDER (wo $\mathcal{L}_{cf}$)         & 57.26/3.43 & 24.69/0.78 & 1.40/0.04 & 80.65/2.34 & 36.88/0.44 & 2.03/0.01 & 141.14/5.78 & 55.48/1.93 & 2.85/0.12 \\ 
CIDER                & 50.56/6.13 & 23.15/1.75 & 1.27/0.09 & 77.62/1.39 & \underline{35.72/0.58} & \underline{1.86/0.05} & 138.53/11.93 & 54.85/3.67 & 2.72/0.14 \\ \hline
Ours (Na\"ive)$\dag$ & 35.87/1.05 & 18.57/0.30 & 1.00/0.01 & 72.68/0.76 & 34.47/0.37 & 1.76/0.03 & 132.28/2.82 & 52.96/0.78 & 2.58/0.05 \\ 
Ours (wo $\mathcal{LLM}$)$\dag$   &35.14/0.94 & 18.56/0.29 & 1.00/0.02 & 72.96/3.05 & 34.29/1.03 & 1.73/0.04 & 130.34/2.77 & \textbf{52.70/0.88} & 2.62/0.06 \\
Ours (wo $\mathcal{R}$)$\dag$     &35.17/0.88 & 18.75/0.34 & 1.00/0.02 & 72.01/2.67 & 34.77/0.87 & 1.75/0.04 & 130.47/2.35 & 53.07/1.34 & \textbf{2.56/0.07} \\
Ours (wo $\mathcal{L}_{cf}$)$\dag$        & 33.72/0.82 & 18.02/0.22 & \textbf{0.99/0.01} & 72.74/2.15 & 34.45/0.48 & 1.77/0.02 & 137.50/1.65 & 55.54/0.65 & 2.74/0.05 \\ 
Ours                 & \textbf{33.31/0.53} & \textbf{18.01/0.29} & \textbf{0.99/0.01} & \textbf{68.91/3.34} & \textbf{33.12/0.91} & \textbf{1.70/0.04} & \textbf{128.92/6.95} & 54.42/1.24 & 2.74/0.07 \\ \hline

Improvements          & 10.43\% & 13.25\% & 10.00\% & 5.80\% & 7.28\% & 9.41\% & 3.65\% & 3.76\% & 7.25\%\\ \hline
\end{tabular}
\label{results}
\end{table*}
\\
\subsection{Experimental Settings}
Our model is developed using PyTorch 1.8.2 and Python 3.8, and is trained on a server equipped with four NVIDIA RTX A6000 GPUs (48GB memory) running the Ubuntu 20.04 operating system. The raw GPS data is interpolated at 15-minute intervals and aggregated into one-hour intervals. For training, we randomly select 80\% of the data for training, 10\% for validation, and 10\% for testing. For POI data categorization, we aggregate the data into 17 categories based on their labels. The input and output windows for the experiments are configured differently, as discussed in the following section. We use the Adam optimizer with an initial learning rate of 0.001 and select the model with the best validation performance using an early stopping strategy before testing. The weight $\alpha$ is set to 1, and we repeat our experiments five times with fixed random seeds. For LLMs, we used LLama3-70b with Ollama, benefiting from its optimizations. This configuration allows the LLM to run with approximately 44GB of GPU memory usage, evenly distributed across the four graphics cards.

\subsection{Overall Performance}\label{performance}
To evaluate the prediction performance of our proposed model, we select several state-of-the-art models in human mobility prediction. It's important to note that it is difficult to implement the event-based treatments into traditional causal-based models, as they are not one-dimensional and cannot be easily classified into binary treatments as in previous studies. Therefore, we mainly focus on the following human mobility prediction models:\\
\textbf{3DGCN} \cite{xia20213dgcn}: A crowd flow prediction model based on a three-dimensional GCN to capture spatio-temporal dependencies.\\
\textbf{GTS} \cite{shang2021discrete}: A time-series prediction model that explores correlation and causation among variables based on GNNs by learning a probabilistic graph by optimizing performance over graph distribution.\\
\textbf{PromptST} \cite{zhang2023promptst}: A spatio-temporal prediction model combining transformers and spatio-temporal prompts to balance common knowledge learning with efficient adaptation to specific tasks. \\
\textbf{STGNN} \cite{wang2020traffic}: A traffic flow prediction model based on GCN, GRU, and transformer structure to learn the latent patterns and capture spatial dependency in traffic nodes of transportation networks.\\
\textbf{DeepST} \cite{zhang2017deep}: A crowd flow prediction model based on ST-ResNET by simulating close, period and trend properties of crowd traffic and further combined with external factors during prediction.\\
\textbf{DeepSTN+} \cite{lin2019deepstn+}: A crowd flow prediction model by modeling the long-range spatial dependence among crowd flows in different regions and fusing POI semantic features simultaneously.\\
\textbf{STID} \cite{shao2022spatial}: A simple MLP-based baseline model for multivariate time series forecasting using spatial and temporal identity embeddings to improve efficiency and accuracy.\\
\textbf{CIDER} \cite{ma2022assessing}: A causal effect estimation framework for COVID-19 policy in the US based on GRU and GCN. Also, Zhang \cite{zhiwen2023assessing} used the same backbone model for assessing causal effects of typhoons.\\
\indent For prediction settings, we assume that public events have different potential effects at multiple time scales, so we design three tasks with different input (In) and output window (Out). We choose Root Mean Square Error (RMSE), Mean Absolute Error (MAE), and Mean Absolute Percentage Error (MAPE) as metrics for comparison.\\
\indent Table \ref{results} illustrates the prediction performance of baseline models and our proposed CausalMob. Overall, our model achieves the best performance in three prediction tasks with repeated experiments using different random seed initializations (see Appendix \ref{hyper}). As the prediction window increases, there is a slight decrease in robustness in standard deviation, but this is acceptable given the excellent prediction accuracy. Specifically, for prediction performances of long-term prediction tasks, although the results show the weakness of our model in both the MAE and MAPE metrics, we believe that the strength of the RMSE is more representative of the outperforming of our model since we use weighted MSE, denoted as $\mathcal{L}_{f}$ in our target function. Additionally, compared with the CIDER which has a different backbone, the results show that our method improves the prediction performances by effectively learning confounders.\\
\indent \textbf{Ablation study} is shown in Table \ref{results} in the last five lines. We evaluate the performances of the na\"ive model (Na\"ive) with simply output $\hat{y}$, the model that replaces all LLMs-derived human intentions with all-zero scores (wo $\mathcal{LLM}$), the model without reweighting mechanism (wo $\mathcal{R}$), the model without counterfactual (wo $\mathcal{L}_{cf}$) loss. We prove that 1) combining public event-based causal inference for human mobility prediction improves prediction performance, and 2) a loss function based on counterfactual inference can further optimize this performance. Similarly, we conduct this ablation study with the CIDER backbone based on Na\"ive and wo $\mathcal{L}_{cf}$, and the results also support the two conclusions we summarized.
\subsection{Analysis of LLMs-derived Human Intentions}\label{humanintentions}
\begin{figure}[ht]
\begin{center}
\includegraphics[scale=0.3]{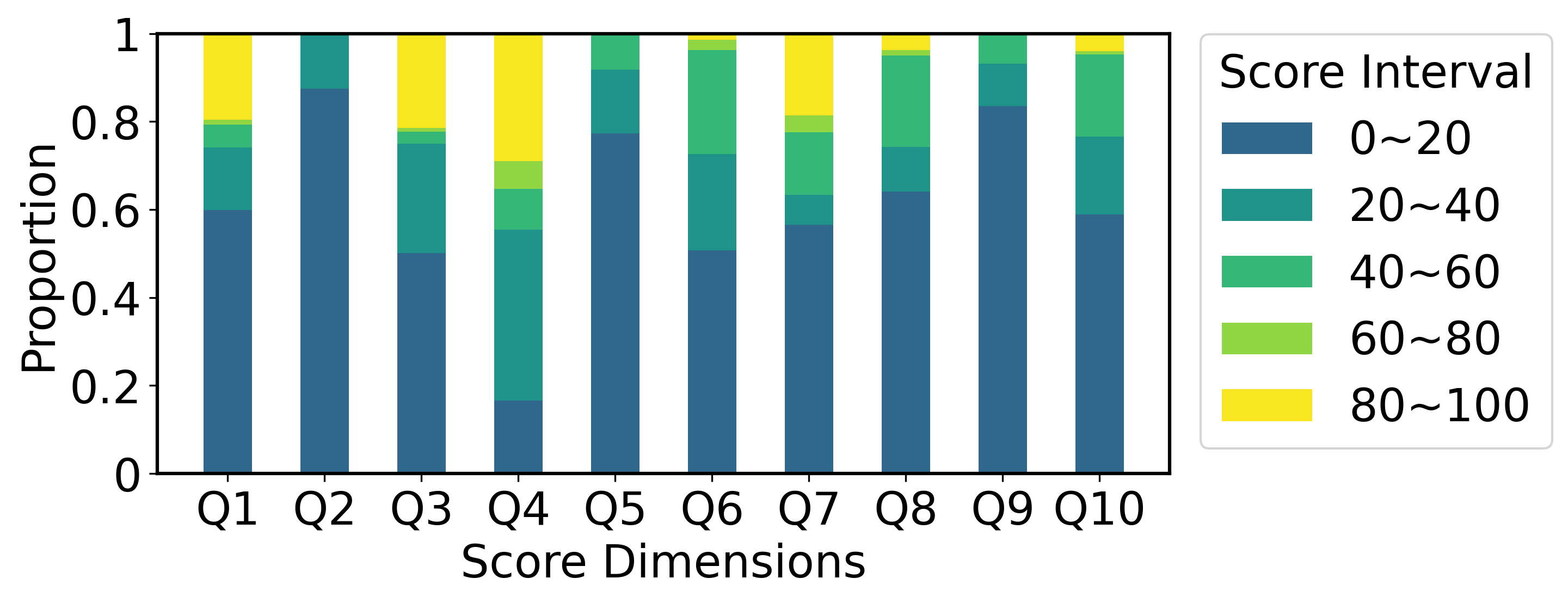}
\end{center}
\caption{Distribution of LLMs-derived human intentions based on scores with prompt engineering. Q1-Q10 refer to 10 questions listed in Prompt 5 of Appendix \ref{prompt}.}
\label{scores}
\end{figure}
\noindent In this section, we analyze LLMs-derived human intentions extracted from structured information. In the previous sections, we set multiple scores to evaluate news articles with Prompt 5 in Appendix \ref{prompt}. In Figure \ref{scores}, we show the distributions of scores for all news articles. The score of Q1 describes the degree of danger of public events, with almost 60\% having no negative impact on human mobility. The third score for Q3 refers to the level of interest, with almost half of the events not attracting people to certain areas. The concepts of danger and interest are not mutually exclusive, allowing an event to receive non-zero scores in both dimensions simultaneously. The remaining scores represent different aspects such as staying in Q2, impact on daily life in Q4, impact on business activities in Q5, transportation in Q6, public health in Q7, government intervention in Q8, and public services in Q9. The last score, Q10, indicates the duration of the public events. With Prompt 4 in Appendix \ref{prompt}, we expect LLMs to judge the predictability of public events. Combined with scores in \textit{danger} and \textit{interest}, we categorize news articles into four quadrants based on public events categorization in Figure \ref{concept}. Finally, public events in news articles labeled with "Yes" are treated as unpredictable cases.\\
\begin{figure}[htbp]
\begin{center}
\includegraphics[scale=0.3]{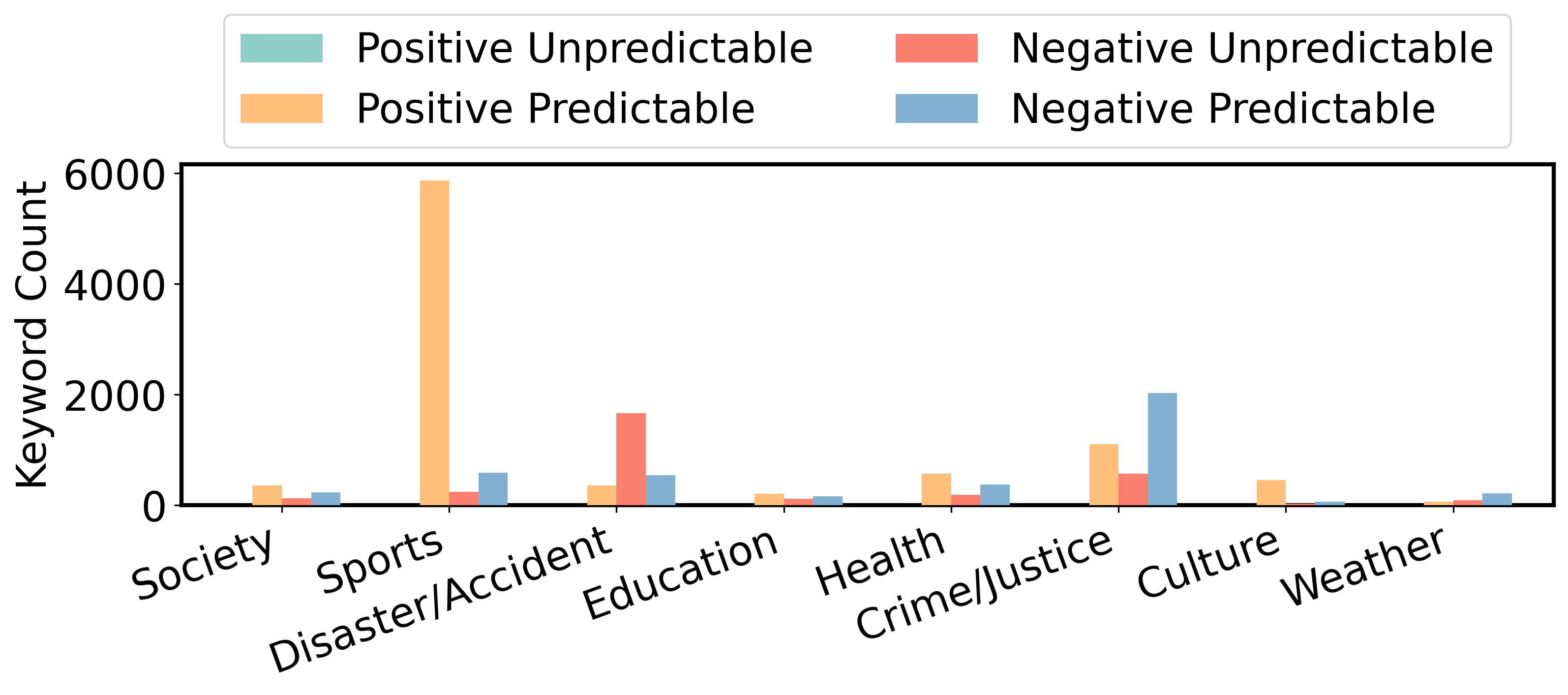}
\end{center}
\caption{Count of news articles labels with public events categorization based on LLMs-derived human intentions}
\label{humanvalue}
\end{figure}
\indent In Figure \ref{humanvalue}, we show the relationships between categorizations based on LLM-derived intentions and labels that the original articles carried. We find that positive unpredictable public events are difficult to extract from our dataset, which is consistent with our considerations for not considering them. Second, positive predictable events comprise the majority of our dataset, especially Sports, Crime/Justice and Cultural related events. Negative events are less represented but dominate the category of Disaster/Accident and are mostly unpredictable. Additionally, negative predictable events are also found in the Crime/Justice. Overall, LLMs can effectively summarize human intentions toward public events from news articles, which is helpful in the analysis of causal inference.
\subsection{Analysis of Encoded Treatment Interpretability}\label{encodevis}
\begin{figure}[htbp]
    \centering
    \includegraphics[width=0.45\textwidth]{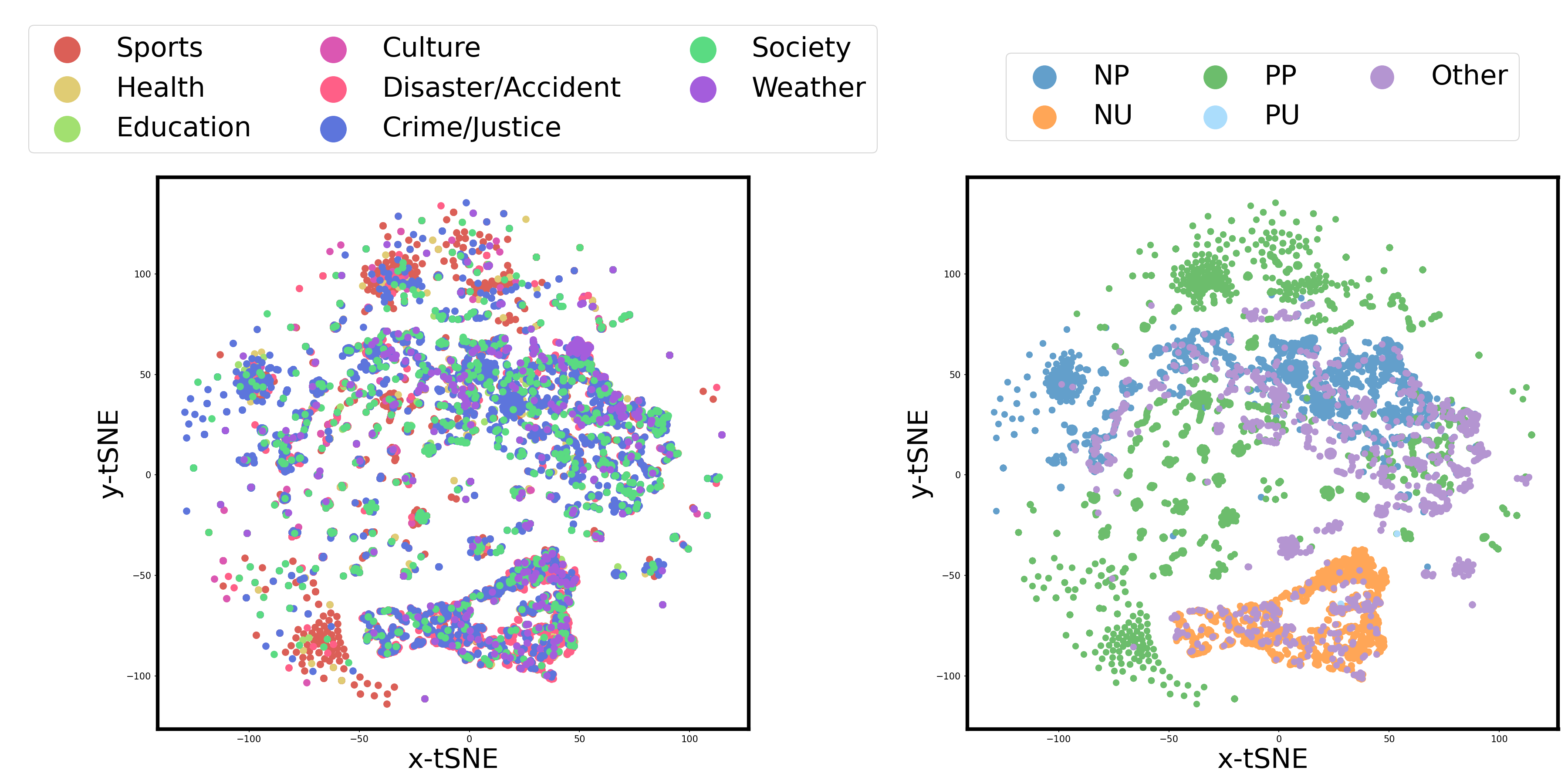}
    \caption{Visualization of encoded treatments with t-SNE. Two figures share the same features, the left figure labeled with news articles types and the right figure is categorized based on human intentions (1st  letter N/P: Negative/Positive, 2nd letter P/U:Predictable/Unpredictable).}
    \label{tsne}
\end{figure}
\noindent In this section, we present the results of encoded human intentions. In our CausalMob, human intentions toward public events in the input and output window are considered simultaneously. All intentions in the input window are included, but only those of predictable events are considered and put into the encoder during training. Here, we choose the encoder in the long-term prediction task to process features obtained from LLMs. In Figure \ref{tsne}, we visualize encoded features of different public events using t-SNE \cite{van2008visualizing}.\\
\indent In Figure.~\ref{tsne}, human intentions features related to Sports are generally distinguished from events with other labels, and the corresponding points in the right figure are categorized into positive predictable events. In contrast, Crime/Justice, Disaster/Accident events have similar distributions. Besides, those corresponding points are always classified into negative events in the right figure. The above conclusions are consistent with the information revealed in Figure \ref{humanvalue}, which also demonstrates the ability of the treatment encoder to identify and learn the characteristics of human intentions.
\subsection{Analysis of Regional Average Treatment Effects}\label{caseeffect}
\begin{figure}[H]
\begin{center}
\includegraphics[scale=0.4]{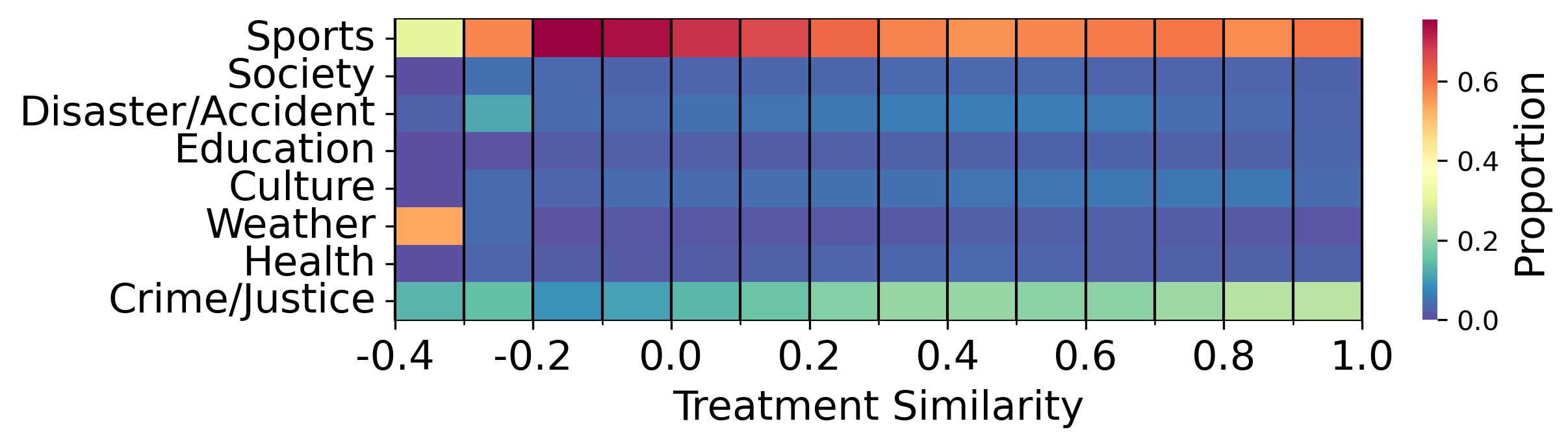}
\end{center}
\caption{Distribution of news articles labels based on treatment similarity compared with treatment baseline.}
\label{treatment_vis}
\end{figure}

\begin{figure}[htbp]
\begin{center}
\includegraphics[scale=0.35]{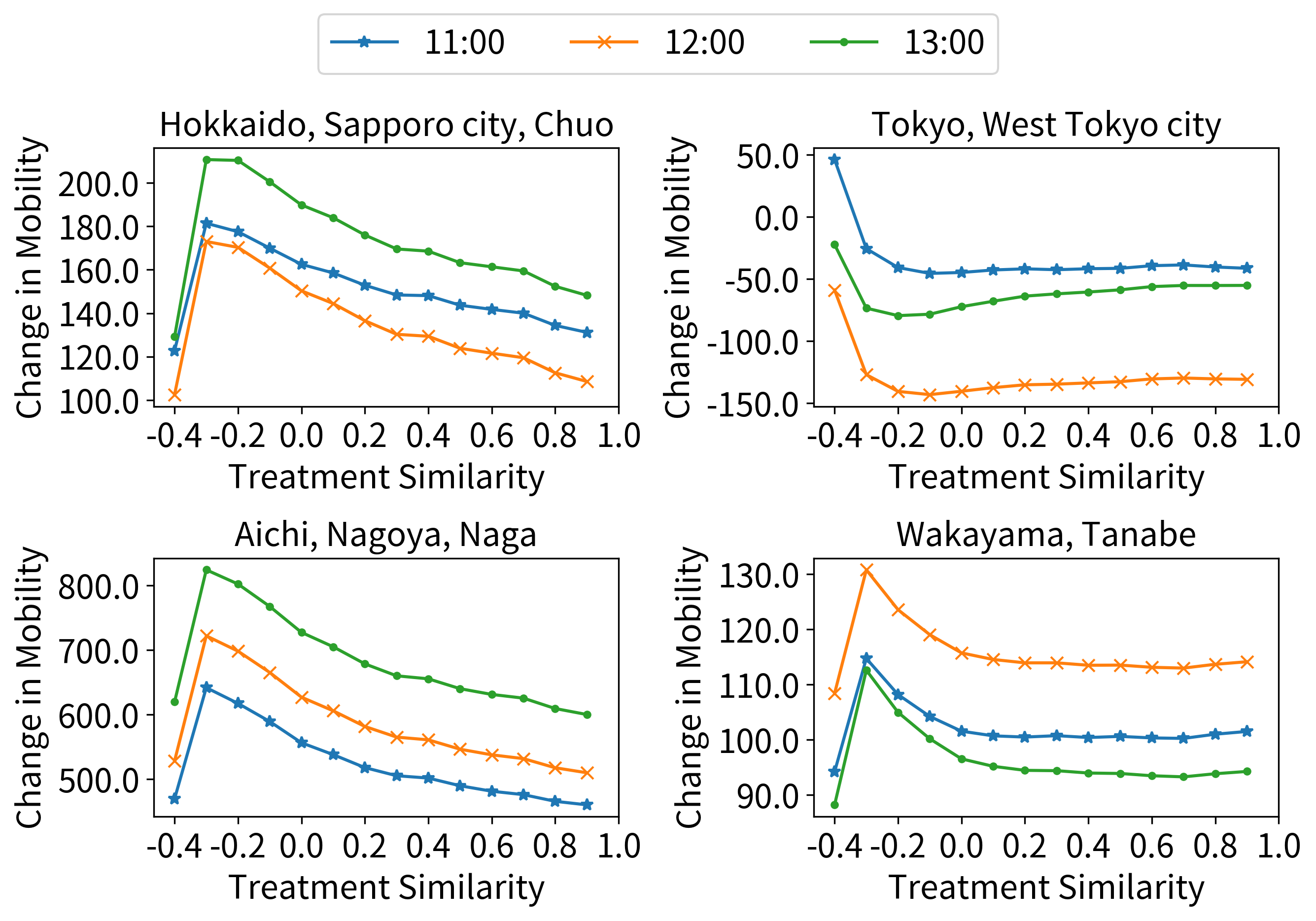}
\end{center}
\caption{Regional average treatment effects of human mobility in multiple prediction timestamps. We selected several regions and predicted average changes of human mobility with treatments in each similarity interval.}
\label{adrf}
\end{figure}

In this section, we analyze the average treatment effects of different public events under the same spatio-temporal conditions. We collect all treatment representations of human intentions during the training process and sorted them according to their cosine similarities to the treatment baseline \cite{singh2021text}. We use 0.1 as the interval to split similarities, resulting in 14 intervals of treatment similarities. In Figure \ref{treatment_vis}, we visualize the distributions of the original news labels in each interval, and we can see that public events related to Sport and Crime/Justice occupy the majority of all treatments intervals. In the most dissimilar treatment interval of $[-0.4, -0.3]$, Weather shares a higher proportion, which is quite different from other intervals.\\
\indent Then, we choose a specific weekday in the test dataset to evaluate the performance with different treatments. In Figure \ref{adrf}, we present the prediction results in selected regions along with the treatment similarity intervals. When treatment is in the most dissimilar interval, as mentioned in Section \ref{caseeffect}, the results show a significant decline in predicted human mobility. As treatment similarity increases towards 1, the change in human mobility gradually decreases (in West Tokyo City, it slightly increases to 0). As we show a similar distribution that constantly changes in Figure \ref{treatment_vis} from -0.3 to 1.0, we also observe a continuous change in human mobility in Figure \ref{adrf}. This demonstrates that our model can control prediction results according to the similarity between encoded treatments of public events and treatment baseline (regular patterns).
\subsection{Analysis of Causal Response with Counterfactual}\label{caseevents}
\begin{figure}[ht]
\begin{center}
\includegraphics[scale=0.45]{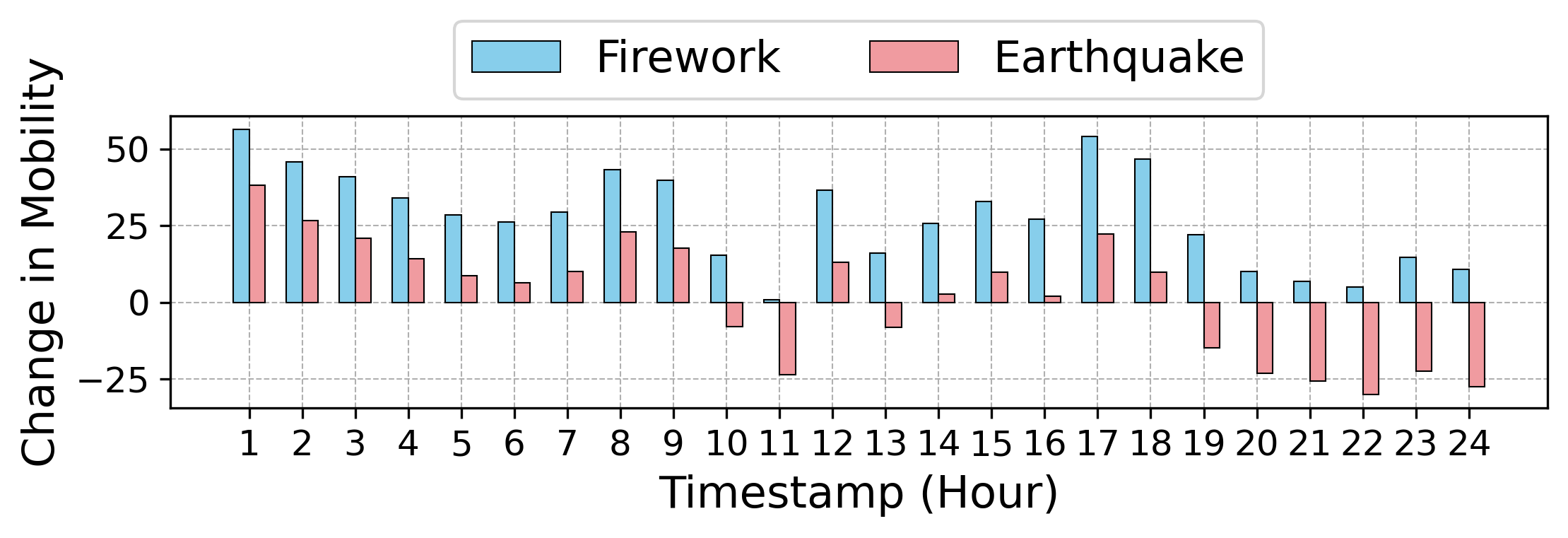}
\end{center}
\caption{Causal Responses with Counterfactual. We show causal responses of human mobility to two public events.}
\label{casepublicevents}
\end{figure}

In this section, we provide an analysis of the causal response of human mobility to different public events as counterfactuals. We select news articles about a firework and an earthquake, then extract structured information to get human intentions as treatments and input them into the long-term prediction process as a predictable event and an unpredictable one, respectively, for one region (Aichi, Matsuyama). In Figure \ref{casepublicevents}, the event feature of the firework festival presents a positive effect on human mobility at all prediction timestamps. On the contrary, the feature of the earthquake tends to control the increase of human mobility compared to the firework, with our results presenting negative effects on human mobility at several timestamps in the mid-noon and night. This demonstrates that our proposed model can simulate the positive and negative effects of different public events on human mobility prediction.
\subsection{Case Study of Prediction with Human Intentions}\label{casestudy}
\begin{figure}[htbp]
\begin{center}
\includegraphics[scale=0.4]{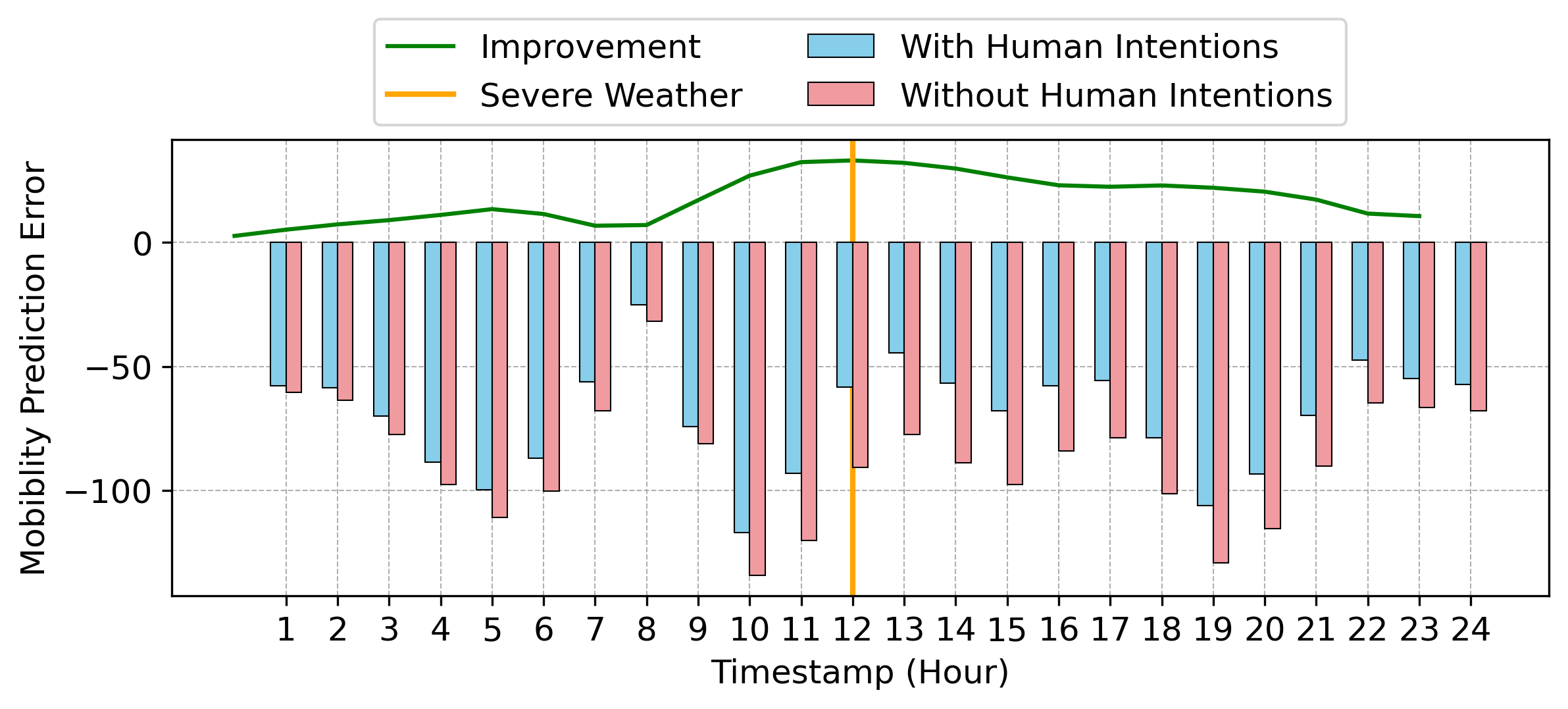}
\end{center}
\caption{Human mobility prediction errors with and without human intentions. The title of the news is as follows:\protect\\Strong winds and heavy rain expected on the Pacific side. Unstable atmospheric conditions due to low pressure system.\protect\\Human Intentions: [80, 20, 0, 90, 40, 60, 70, 80, 30, 60]}
\label{case_study}
\end{figure}

In this section, we provide a case study in test dataset to show the effect of our model. In Figure \ref{case_study}, a predictable public event of severe weather is reported on 2024-03-12 08:25:00, and in the structured information processed with LLMs, the event time is reset to noon the next day. We use LLMs-derived human intentions, as shown in the title of Figure \ref{case_study}, to predict human mobility on the day following the 12th. For comparison, we use all-zero intentions to mask this event and also predict future human mobility. Then, we calculate the prediction errors for each timestamp and the improvement of our model with the green line, along with the occurrence of the severe weather at noon, indicated by the orange line.\\
\indent In Figure \ref{case_study}, we can see that the prediction results are similar in the morning hours of the day. However, as the predictable severe weather approaches, our model starts to show more accurate predictions, which is evident from the change in the Improvement line. In this case, our model achieved the best results around the predictable public event (i.e., 12:00 p.m.). The improved performance of predictions with and without the causal influence of human intentions decreases gradually after that point. Overall, our model consistently maintained more accurate predictions. This result shows that our model can use structured information and human intentions from news articles to enhance human mobility prediction during predictable non-periodical public events.
\section{Conclusion}
In this research, we predict human mobility based on LLM-derived human intentions toward public events using a causal inference framework and evaluate the response of human mobility to different public events. We utilize news articles to extract structured information about public events and generate human intentions with LLMs through prompt engineering. We propose a prediction model called \textbf{CausalMob} to predict human mobility and estimate the causal effect of public events using multiple data sources (including POI data and human mobility data from GPS trajectories) in the main administrative regions of Japan. Compared with several state-of-the-art models, our proposed model performs better on multiple prediction tasks. Furthermore, our approach provides interpretable results for investigating the relationship between human mobility and public events. Our findings can help decision-makers estimate mobility changes in different regions and make informed decisions in advance when faced with various public events.


\bibliographystyle{ACM-Reference-Format}
\balance
\bibliography{sample-base}


\begin{thebibliography}{48}


\ifx \showCODEN    \undefined \def \showCODEN     #1{\unskip}     \fi
\ifx \showDOI      \undefined \def \showDOI       #1{#1}\fi
\ifx \showISBNx    \undefined \def \showISBNx     #1{\unskip}     \fi
\ifx \showISBNxiii \undefined \def \showISBNxiii  #1{\unskip}     \fi
\ifx \showISSN     \undefined \def \showISSN      #1{\unskip}     \fi
\ifx \showLCCN     \undefined \def \showLCCN      #1{\unskip}     \fi
\ifx \shownote     \undefined \def \shownote      #1{#1}          \fi
\ifx \showarticletitle \undefined \def \showarticletitle #1{#1}   \fi
\ifx \showURL      \undefined \def \showURL       {\relax}        \fi
\providecommand\bibfield[2]{#2}
\providecommand\bibinfo[2]{#2}
\providecommand\natexlab[1]{#1}
\providecommand\showeprint[2][]{arXiv:#2}

\bibitem[Blogwatcher(2024)]%
        {blogwatcher}
\bibfield{author}{\bibinfo{person}{Blogwatcher}.} \bibinfo{year}{2024}\natexlab{}.
\newblock \bibinfo{title}{Blogwatcher, Inc.}
\newblock
\urldef\tempurl%
\url{https://www.blogwatcher.co.jp/}
\showURL{%
\tempurl}
\newblock
\shownote{(Japanese Only)}.


\bibitem[Chen et~al\mbox{.}(2020)]%
        {chen2020dualsin}
\bibfield{author}{\bibinfo{person}{Quanjun Chen}, \bibinfo{person}{Renhe Jiang}, \bibinfo{person}{Chuang Yang}, \bibinfo{person}{Zekun Cai}, \bibinfo{person}{Zipei Fan}, \bibinfo{person}{Kota Tsubouchi}, \bibinfo{person}{Ryosuke Shibasaki}, {and} \bibinfo{person}{Xuan Song}.} \bibinfo{year}{2020}\natexlab{}.
\newblock \showarticletitle{Dualsin: Dual sequential interaction network for human intentional mobility prediction}. In \bibinfo{booktitle}{\emph{Proceedings of the 28th International Conference on Advances in Geographic Information Systems}}. \bibinfo{pages}{283--292}.
\newblock


\bibitem[Fan et~al\mbox{.}(2018)]%
        {fan2018online}
\bibfield{author}{\bibinfo{person}{Zipei Fan}, \bibinfo{person}{Xuan Song}, \bibinfo{person}{Tianqi Xia}, \bibinfo{person}{Renhe Jiang}, \bibinfo{person}{Ryosuke Shibasaki}, {and} \bibinfo{person}{Ritsu Sakuramachi}.} \bibinfo{year}{2018}\natexlab{}.
\newblock \showarticletitle{Online deep ensemble learning for predicting citywide human mobility}.
\newblock \bibinfo{journal}{\emph{Proceedings of the ACM on Interactive, Mobile, Wearable and Ubiquitous Technologies}} \bibinfo{volume}{2}, \bibinfo{number}{3} (\bibinfo{year}{2018}), \bibinfo{pages}{1--21}.
\newblock


\bibitem[Farmer et~al\mbox{.}(2018)]%
        {farmer2018application}
\bibfield{author}{\bibinfo{person}{Ruth~E Farmer}, \bibinfo{person}{Daphne Kounali}, \bibinfo{person}{A~Sarah Walker}, \bibinfo{person}{Jelena Savovi{\'c}}, \bibinfo{person}{Alison Richards}, \bibinfo{person}{Margaret~T May}, {and} \bibinfo{person}{Deborah Ford}.} \bibinfo{year}{2018}\natexlab{}.
\newblock \showarticletitle{Application of causal inference methods in the analyses of randomised controlled trials: a systematic review}.
\newblock \bibinfo{journal}{\emph{Trials}}  \bibinfo{volume}{19} (\bibinfo{year}{2018}), \bibinfo{pages}{1--14}.
\newblock


\bibitem[Gretton et~al\mbox{.}(2012)]%
        {gretton2012kernel}
\bibfield{author}{\bibinfo{person}{Arthur Gretton}, \bibinfo{person}{Karsten~M Borgwardt}, \bibinfo{person}{Malte~J Rasch}, \bibinfo{person}{Bernhard Sch{\"o}lkopf}, {and} \bibinfo{person}{Alexander Smola}.} \bibinfo{year}{2012}\natexlab{}.
\newblock \showarticletitle{A kernel two-sample test}.
\newblock \bibinfo{journal}{\emph{The Journal of Machine Learning Research}} \bibinfo{volume}{13}, \bibinfo{number}{1} (\bibinfo{year}{2012}), \bibinfo{pages}{723--773}.
\newblock


\bibitem[Hangli et~al\mbox{.}(2022)]%
        {hangli2022multi}
\bibfield{author}{\bibinfo{person}{Ge Hangli}, \bibinfo{person}{Lifeng Lin}, \bibinfo{person}{Renhe Jiang}, \bibinfo{person}{Takashi Michikata}, {and} \bibinfo{person}{Noboru Koshizuka}.} \bibinfo{year}{2022}\natexlab{}.
\newblock \showarticletitle{Multi-Weighted Graphs Learning for Passenger Count Prediction on Railway Network}. In \bibinfo{booktitle}{\emph{2022 IEEE 46th Annual Computers, Software, and Applications Conference (COMPSAC)}}. IEEE, \bibinfo{pages}{374--382}.
\newblock


\bibitem[Hasan et~al\mbox{.}(2024)]%
        {hasan2024llm}
\bibfield{author}{\bibinfo{person}{Md~Rakibul Hasan}, \bibinfo{person}{Md~Zakir Hossain}, \bibinfo{person}{Tom Gedeon}, {and} \bibinfo{person}{Shafin Rahman}.} \bibinfo{year}{2024}\natexlab{}.
\newblock \showarticletitle{LLM-GEm: Large language model-guided prediction of people’s empathy levels towards newspaper article}. In \bibinfo{booktitle}{\emph{Findings of the Association for Computational Linguistics: EACL 2024}}. \bibinfo{pages}{2215--2231}.
\newblock


\bibitem[Hasan et~al\mbox{.}(2013)]%
        {hasan2013spatiotemporal}
\bibfield{author}{\bibinfo{person}{Samiul Hasan}, \bibinfo{person}{Christian~M Schneider}, \bibinfo{person}{Satish~V Ukkusuri}, {and} \bibinfo{person}{Marta~C Gonz{\'a}lez}.} \bibinfo{year}{2013}\natexlab{}.
\newblock \showarticletitle{Spatiotemporal patterns of urban human mobility}.
\newblock \bibinfo{journal}{\emph{Journal of Statistical Physics}}  \bibinfo{volume}{151} (\bibinfo{year}{2013}), \bibinfo{pages}{304--318}.
\newblock


\bibitem[He et~al\mbox{.}(2016)]%
        {he2016deep}
\bibfield{author}{\bibinfo{person}{Kaiming He}, \bibinfo{person}{Xiangyu Zhang}, \bibinfo{person}{Shaoqing Ren}, {and} \bibinfo{person}{Jian Sun}.} \bibinfo{year}{2016}\natexlab{}.
\newblock \showarticletitle{Deep residual learning for image recognition}. In \bibinfo{booktitle}{\emph{Proceedings of the IEEE conference on computer vision and pattern recognition}}. \bibinfo{pages}{770--778}.
\newblock


\bibitem[Imbens(2000)]%
        {imbens2000role}
\bibfield{author}{\bibinfo{person}{Guido~W Imbens}.} \bibinfo{year}{2000}\natexlab{}.
\newblock \showarticletitle{The role of the propensity score in estimating dose-response functions}.
\newblock \bibinfo{journal}{\emph{Biometrika}} \bibinfo{volume}{87}, \bibinfo{number}{3} (\bibinfo{year}{2000}), \bibinfo{pages}{706--710}.
\newblock


\bibitem[Jiang et~al\mbox{.}(2023b)]%
        {jiang2023structgpt}
\bibfield{author}{\bibinfo{person}{Jinhao Jiang}, \bibinfo{person}{Kun Zhou}, \bibinfo{person}{Zican Dong}, \bibinfo{person}{Keming Ye}, \bibinfo{person}{Wayne~Xin Zhao}, {and} \bibinfo{person}{Ji-Rong Wen}.} \bibinfo{year}{2023}\natexlab{b}.
\newblock \showarticletitle{Structgpt: A general framework for large language model to reason over structured data}.
\newblock \bibinfo{journal}{\emph{arXiv preprint arXiv:2305.09645}} (\bibinfo{year}{2023}).
\newblock


\bibitem[Jiang et~al\mbox{.}(2022)]%
        {jiang2022will}
\bibfield{author}{\bibinfo{person}{Renhe Jiang}, \bibinfo{person}{Quanjun Chen}, \bibinfo{person}{Zekun Cai}, \bibinfo{person}{Zipei Fan}, \bibinfo{person}{Xuan Song}, \bibinfo{person}{Kota Tsubouchi}, {and} \bibinfo{person}{Ryosuke Shibasaki}.} \bibinfo{year}{2022}\natexlab{}.
\newblock \showarticletitle{Will you go where you search? A deep learning framework for estimating user search-and-go behavior}.
\newblock \bibinfo{journal}{\emph{Neurocomputing}}  \bibinfo{volume}{472} (\bibinfo{year}{2022}), \bibinfo{pages}{338--348}.
\newblock


\bibitem[Jiang et~al\mbox{.}(2019)]%
        {jiang2019deepurbanevent}
\bibfield{author}{\bibinfo{person}{Renhe Jiang}, \bibinfo{person}{Xuan Song}, \bibinfo{person}{Dou Huang}, \bibinfo{person}{Xiaoya Song}, \bibinfo{person}{Tianqi Xia}, \bibinfo{person}{Zekun Cai}, \bibinfo{person}{Zhaonan Wang}, \bibinfo{person}{Kyoung-Sook Kim}, {and} \bibinfo{person}{Ryosuke Shibasaki}.} \bibinfo{year}{2019}\natexlab{}.
\newblock \showarticletitle{Deepurbanevent: A system for predicting citywide crowd dynamics at big events}. In \bibinfo{booktitle}{\emph{Proceedings of the 25th ACM SIGKDD international conference on knowledge discovery \& data mining}}. \bibinfo{pages}{2114--2122}.
\newblock


\bibitem[Jiang et~al\mbox{.}(2023a)]%
        {jiang2023learning}
\bibfield{author}{\bibinfo{person}{Renhe Jiang}, \bibinfo{person}{Zhaonan Wang}, \bibinfo{person}{Yudong Tao}, \bibinfo{person}{Chuang Yang}, \bibinfo{person}{Xuan Song}, \bibinfo{person}{Ryosuke Shibasaki}, \bibinfo{person}{Shu-Ching Chen}, {and} \bibinfo{person}{Mei-Ling Shyu}.} \bibinfo{year}{2023}\natexlab{a}.
\newblock \showarticletitle{Learning social meta-knowledge for nowcasting human mobility in disaster}. In \bibinfo{booktitle}{\emph{Proceedings of the ACM Web Conference 2023}}. \bibinfo{pages}{2655--2665}.
\newblock


\bibitem[Jin et~al\mbox{.}(2024)]%
        {jin2024comprehensive}
\bibfield{author}{\bibinfo{person}{Hanlei Jin}, \bibinfo{person}{Yang Zhang}, \bibinfo{person}{Dan Meng}, \bibinfo{person}{Jun Wang}, {and} \bibinfo{person}{Jinghua Tan}.} \bibinfo{year}{2024}\natexlab{}.
\newblock \showarticletitle{A comprehensive survey on process-oriented automatic text summarization with exploration of llm-based methods}.
\newblock \bibinfo{journal}{\emph{arXiv preprint arXiv:2403.02901}} (\bibinfo{year}{2024}).
\newblock


\bibitem[Kipf and Welling(2016)]%
        {kipf2016semi}
\bibfield{author}{\bibinfo{person}{Thomas~N Kipf} {and} \bibinfo{person}{Max Welling}.} \bibinfo{year}{2016}\natexlab{}.
\newblock \showarticletitle{Semi-supervised classification with graph convolutional networks}.
\newblock \bibinfo{journal}{\emph{arXiv preprint arXiv:1609.02907}} (\bibinfo{year}{2016}).
\newblock


\bibitem[Li et~al\mbox{.}(2024)]%
        {li2024causalstock}
\bibfield{author}{\bibinfo{person}{Shuqi Li}, \bibinfo{person}{Yuebo Sun}, \bibinfo{person}{Yuxin Lin}, \bibinfo{person}{Xin Gao}, \bibinfo{person}{Shuo Shang}, {and} \bibinfo{person}{Rui Yan}.} \bibinfo{year}{2024}\natexlab{}.
\newblock \showarticletitle{CausalStock: Deep End-to-end Causal Discovery for News-driven Stock Movement Prediction}.
\newblock \bibinfo{journal}{\emph{arXiv preprint arXiv:2411.06391}} (\bibinfo{year}{2024}).
\newblock


\bibitem[Li et~al\mbox{.}(2023)]%
        {li2023metaagents}
\bibfield{author}{\bibinfo{person}{Yuan Li}, \bibinfo{person}{Yixuan Zhang}, {and} \bibinfo{person}{Lichao Sun}.} \bibinfo{year}{2023}\natexlab{}.
\newblock \showarticletitle{Metaagents: Simulating interactions of human behaviors for llm-based task-oriented coordination via collaborative generative agents}.
\newblock \bibinfo{journal}{\emph{arXiv preprint arXiv:2310.06500}} (\bibinfo{year}{2023}).
\newblock


\bibitem[Lin et~al\mbox{.}(2019)]%
        {lin2019deepstn+}
\bibfield{author}{\bibinfo{person}{Ziqian Lin}, \bibinfo{person}{Jie Feng}, \bibinfo{person}{Ziyang Lu}, \bibinfo{person}{Yong Li}, {and} \bibinfo{person}{Depeng Jin}.} \bibinfo{year}{2019}\natexlab{}.
\newblock \showarticletitle{Deepstn+: Context-aware spatial-temporal neural network for crowd flow prediction in metropolis}. In \bibinfo{booktitle}{\emph{Proceedings of the AAAI conference on artificial intelligence}}, Vol.~\bibinfo{volume}{33}. \bibinfo{pages}{1020--1027}.
\newblock


\bibitem[Ma et~al\mbox{.}(2022)]%
        {ma2022assessing}
\bibfield{author}{\bibinfo{person}{Jing Ma}, \bibinfo{person}{Yushun Dong}, \bibinfo{person}{Zheng Huang}, \bibinfo{person}{Daniel Mietchen}, {and} \bibinfo{person}{Jundong Li}.} \bibinfo{year}{2022}\natexlab{}.
\newblock \showarticletitle{Assessing the causal impact of COVID-19 related policies on outbreak dynamics: A case study in the US}. In \bibinfo{booktitle}{\emph{Proceedings of the ACM Web Conference 2022}}. \bibinfo{pages}{2678--2686}.
\newblock


\bibitem[Morgan and Winship(2015)]%
        {morgan2015counterfactuals}
\bibfield{author}{\bibinfo{person}{Stephen~L Morgan} {and} \bibinfo{person}{Christopher Winship}.} \bibinfo{year}{2015}\natexlab{}.
\newblock \bibinfo{booktitle}{\emph{Counterfactuals and causal inference}}.
\newblock \bibinfo{publisher}{Cambridge University Press}.
\newblock


\bibitem[ollama(2024)]%
        {repository}
\bibfield{author}{\bibinfo{person}{ollama}.} \bibinfo{year}{2024}\natexlab{}.
\newblock \bibinfo{title}{ollama}.
\newblock
\urldef\tempurl%
\url{https://github.com/ollama/ollama}
\showURL{%
\tempurl}


\bibitem[Pearl(2009)]%
        {pearl2009causal}
\bibfield{author}{\bibinfo{person}{Judea Pearl}.} \bibinfo{year}{2009}\natexlab{}.
\newblock \showarticletitle{Causal inference in statistics: An overview}.
\newblock  (\bibinfo{year}{2009}).
\newblock


\bibitem[Runfeng et~al\mbox{.}(2023)]%
        {runfeng2023lkpnr}
\bibfield{author}{\bibinfo{person}{Xie Runfeng}, \bibinfo{person}{Cui Xiangyang}, \bibinfo{person}{Yan Zhou}, \bibinfo{person}{Wang Xin}, \bibinfo{person}{Xuan Zhanwei}, \bibinfo{person}{Zhang Kai}, {et~al\mbox{.}}} \bibinfo{year}{2023}\natexlab{}.
\newblock \showarticletitle{Lkpnr: Llm and kg for personalized news recommendation framework}.
\newblock \bibinfo{journal}{\emph{arXiv preprint arXiv:2308.12028}} (\bibinfo{year}{2023}).
\newblock


\bibitem[Schwab et~al\mbox{.}(2020)]%
        {schwab2020learning}
\bibfield{author}{\bibinfo{person}{Patrick Schwab}, \bibinfo{person}{Lorenz Linhardt}, \bibinfo{person}{Stefan Bauer}, \bibinfo{person}{Joachim~M Buhmann}, {and} \bibinfo{person}{Walter Karlen}.} \bibinfo{year}{2020}\natexlab{}.
\newblock \showarticletitle{Learning counterfactual representations for estimating individual dose-response curves}. In \bibinfo{booktitle}{\emph{Proceedings of the AAAI Conference on Artificial Intelligence}}, Vol.~\bibinfo{volume}{34}. \bibinfo{pages}{5612--5619}.
\newblock


\bibitem[Shang et~al\mbox{.}(2021)]%
        {shang2021discrete}
\bibfield{author}{\bibinfo{person}{Chao Shang}, \bibinfo{person}{Jie Chen}, {and} \bibinfo{person}{Jinbo Bi}.} \bibinfo{year}{2021}\natexlab{}.
\newblock \showarticletitle{Discrete graph structure learning for forecasting multiple time series}.
\newblock \bibinfo{journal}{\emph{arXiv preprint arXiv:2101.06861}} (\bibinfo{year}{2021}).
\newblock


\bibitem[Shao et~al\mbox{.}(2022)]%
        {shao2022spatial}
\bibfield{author}{\bibinfo{person}{Zezhi Shao}, \bibinfo{person}{Zhao Zhang}, \bibinfo{person}{Fei Wang}, \bibinfo{person}{Wei Wei}, {and} \bibinfo{person}{Yongjun Xu}.} \bibinfo{year}{2022}\natexlab{}.
\newblock \showarticletitle{Spatial-temporal identity: A simple yet effective baseline for multivariate time series forecasting}. In \bibinfo{booktitle}{\emph{Proceedings of the 31st ACM International Conference on Information \& Knowledge Management}}. \bibinfo{pages}{4454--4458}.
\newblock


\bibitem[Singh and Singh(2021)]%
        {singh2021text}
\bibfield{author}{\bibinfo{person}{Ritika Singh} {and} \bibinfo{person}{Satwinder Singh}.} \bibinfo{year}{2021}\natexlab{}.
\newblock \showarticletitle{Text similarity measures in news articles by vector space model using NLP}.
\newblock \bibinfo{journal}{\emph{Journal of The Institution of Engineers (India): Series B}}  \bibinfo{volume}{102} (\bibinfo{year}{2021}), \bibinfo{pages}{329--338}.
\newblock


\bibitem[Song et~al\mbox{.}(2010)]%
        {song2010limits}
\bibfield{author}{\bibinfo{person}{Chaoming Song}, \bibinfo{person}{Zehui Qu}, \bibinfo{person}{Nicholas Blumm}, {and} \bibinfo{person}{Albert-L{\'a}szl{\'o} Barab{\'a}si}.} \bibinfo{year}{2010}\natexlab{}.
\newblock \showarticletitle{Limits of predictability in human mobility}.
\newblock \bibinfo{journal}{\emph{Science}} \bibinfo{volume}{327}, \bibinfo{number}{5968} (\bibinfo{year}{2010}), \bibinfo{pages}{1018--1021}.
\newblock


\bibitem[Song et~al\mbox{.}(2016)]%
        {song2016prediction}
\bibfield{author}{\bibinfo{person}{Xuan Song}, \bibinfo{person}{Quanshi Zhang}, \bibinfo{person}{Yoshihide Sekimoto}, \bibinfo{person}{Ryosuke Shibasaki}, \bibinfo{person}{Nicholas~Jing Yuan}, {and} \bibinfo{person}{Xing Xie}.} \bibinfo{year}{2016}\natexlab{}.
\newblock \showarticletitle{Prediction and simulation of human mobility following natural disasters}.
\newblock \bibinfo{journal}{\emph{ACM Transactions on Intelligent Systems and Technology (TIST)}} \bibinfo{volume}{8}, \bibinfo{number}{2} (\bibinfo{year}{2016}), \bibinfo{pages}{1--23}.
\newblock


\bibitem[Touvron et~al\mbox{.}(2023)]%
        {touvron2023llama}
\bibfield{author}{\bibinfo{person}{Hugo Touvron}, \bibinfo{person}{Thibaut Lavril}, \bibinfo{person}{Gautier Izacard}, \bibinfo{person}{Xavier Martinet}, \bibinfo{person}{Marie-Anne Lachaux}, \bibinfo{person}{Timoth{\'e}e Lacroix}, \bibinfo{person}{Baptiste Rozi{\`e}re}, \bibinfo{person}{Naman Goyal}, \bibinfo{person}{Eric Hambro}, \bibinfo{person}{Faisal Azhar}, {et~al\mbox{.}}} \bibinfo{year}{2023}\natexlab{}.
\newblock \showarticletitle{Llama: Open and efficient foundation language models}.
\newblock \bibinfo{journal}{\emph{arXiv preprint arXiv:2302.13971}} (\bibinfo{year}{2023}).
\newblock


\bibitem[Van~der Maaten and Hinton(2008)]%
        {van2008visualizing}
\bibfield{author}{\bibinfo{person}{Laurens Van~der Maaten} {and} \bibinfo{person}{Geoffrey Hinton}.} \bibinfo{year}{2008}\natexlab{}.
\newblock \showarticletitle{Visualizing data using t-SNE.}
\newblock \bibinfo{journal}{\emph{Journal of machine learning research}} \bibinfo{volume}{9}, \bibinfo{number}{11} (\bibinfo{year}{2008}).
\newblock


\bibitem[Vijayan(2023)]%
        {vijayan2023prompt}
\bibfield{author}{\bibinfo{person}{Aishwarya Vijayan}.} \bibinfo{year}{2023}\natexlab{}.
\newblock \showarticletitle{A prompt engineering approach for structured data extraction from unstructured text using conversational LLMs}. In \bibinfo{booktitle}{\emph{Proceedings of the 2023 6th International Conference on Algorithms, Computing and Artificial Intelligence}}. \bibinfo{pages}{183--189}.
\newblock


\bibitem[Wang et~al\mbox{.}(2023)]%
        {wang2023would}
\bibfield{author}{\bibinfo{person}{Xinglei Wang}, \bibinfo{person}{Meng Fang}, \bibinfo{person}{Zichao Zeng}, {and} \bibinfo{person}{Tao Cheng}.} \bibinfo{year}{2023}\natexlab{}.
\newblock \showarticletitle{Where would i go next? large language models as human mobility predictors}.
\newblock \bibinfo{journal}{\emph{arXiv preprint arXiv:2308.15197}} (\bibinfo{year}{2023}).
\newblock


\bibitem[Wang et~al\mbox{.}(2024)]%
        {wang2024news}
\bibfield{author}{\bibinfo{person}{Xinlei Wang}, \bibinfo{person}{Maike Feng}, \bibinfo{person}{Jing Qiu}, \bibinfo{person}{Jinjin Gu}, {and} \bibinfo{person}{Junhua Zhao}.} \bibinfo{year}{2024}\natexlab{}.
\newblock \showarticletitle{From news to forecast: Integrating event analysis in llm-based time series forecasting with reflection}.
\newblock \bibinfo{journal}{\emph{arXiv preprint arXiv:2409.17515}} (\bibinfo{year}{2024}).
\newblock


\bibitem[Wang et~al\mbox{.}(2022b)]%
        {wang2022generalization}
\bibfield{author}{\bibinfo{person}{Xin Wang}, \bibinfo{person}{Shengfei Lyu}, \bibinfo{person}{Xingyu Wu}, \bibinfo{person}{Tianhao Wu}, {and} \bibinfo{person}{Huanhuan Chen}.} \bibinfo{year}{2022}\natexlab{b}.
\newblock \showarticletitle{Generalization bounds for estimating causal effects of continuous treatments}.
\newblock \bibinfo{journal}{\emph{Advances in Neural Information Processing Systems}}  \bibinfo{volume}{35} (\bibinfo{year}{2022}), \bibinfo{pages}{8605--8617}.
\newblock


\bibitem[Wang et~al\mbox{.}(2020)]%
        {wang2020traffic}
\bibfield{author}{\bibinfo{person}{Xiaoyang Wang}, \bibinfo{person}{Yao Ma}, \bibinfo{person}{Yiqi Wang}, \bibinfo{person}{Wei Jin}, \bibinfo{person}{Xin Wang}, \bibinfo{person}{Jiliang Tang}, \bibinfo{person}{Caiyan Jia}, {and} \bibinfo{person}{Jian Yu}.} \bibinfo{year}{2020}\natexlab{}.
\newblock \showarticletitle{Traffic flow prediction via spatial temporal graph neural network}. In \bibinfo{booktitle}{\emph{Proceedings of the web conference 2020}}. \bibinfo{pages}{1082--1092}.
\newblock


\bibitem[Wang et~al\mbox{.}(2015)]%
        {wang2015regularity}
\bibfield{author}{\bibinfo{person}{Yingzi Wang}, \bibinfo{person}{Nicholas~Jing Yuan}, \bibinfo{person}{Defu Lian}, \bibinfo{person}{Linli Xu}, \bibinfo{person}{Xing Xie}, \bibinfo{person}{Enhong Chen}, {and} \bibinfo{person}{Yong Rui}.} \bibinfo{year}{2015}\natexlab{}.
\newblock \showarticletitle{Regularity and conformity: Location prediction using heterogeneous mobility data}. In \bibinfo{booktitle}{\emph{Proceedings of the 21th ACM SIGKDD international conference on knowledge discovery and data mining}}. \bibinfo{pages}{1275--1284}.
\newblock


\bibitem[Wang et~al\mbox{.}(2022a)]%
        {wang2022event}
\bibfield{author}{\bibinfo{person}{Zhaonan Wang}, \bibinfo{person}{Renhe Jiang}, \bibinfo{person}{Hao Xue}, \bibinfo{person}{Flora~D Salim}, \bibinfo{person}{Xuan Song}, {and} \bibinfo{person}{Ryosuke Shibasaki}.} \bibinfo{year}{2022}\natexlab{a}.
\newblock \showarticletitle{Event-aware multimodal mobility nowcasting}. In \bibinfo{booktitle}{\emph{Proceedings of the AAAI Conference on Artificial Intelligence}}, Vol.~\bibinfo{volume}{36}. \bibinfo{pages}{4228--4236}.
\newblock


\bibitem[Wei et~al\mbox{.}(2022)]%
        {wei2022chain}
\bibfield{author}{\bibinfo{person}{Jason Wei}, \bibinfo{person}{Xuezhi Wang}, \bibinfo{person}{Dale Schuurmans}, \bibinfo{person}{Maarten Bosma}, \bibinfo{person}{Fei Xia}, \bibinfo{person}{Ed Chi}, \bibinfo{person}{Quoc~V Le}, \bibinfo{person}{Denny Zhou}, {et~al\mbox{.}}} \bibinfo{year}{2022}\natexlab{}.
\newblock \showarticletitle{Chain-of-thought prompting elicits reasoning in large language models}.
\newblock \bibinfo{journal}{\emph{Advances in neural information processing systems}}  \bibinfo{volume}{35} (\bibinfo{year}{2022}), \bibinfo{pages}{24824--24837}.
\newblock


\bibitem[{Wikipedia contributors}(2023)]%
        {enwiki:1187452593}
\bibfield{author}{\bibinfo{person}{{Wikipedia contributors}}.} \bibinfo{year}{2023}\natexlab{}.
\newblock \bibinfo{title}{Kyodo News --- {Wikipedia}{,} The Free Encyclopedia}.
\newblock
\urldef\tempurl%
\url{https://en.wikipedia.org/w/index.php?title=Kyodo_News&oldid=1187452593}
\showURL{%
\tempurl}
\newblock
\shownote{[Online; accessed 8-July-2024]}.


\bibitem[Xia et~al\mbox{.}(2021)]%
        {xia20213dgcn}
\bibfield{author}{\bibinfo{person}{Tong Xia}, \bibinfo{person}{Junjie Lin}, \bibinfo{person}{Yong Li}, \bibinfo{person}{Jie Feng}, \bibinfo{person}{Pan Hui}, \bibinfo{person}{Funing Sun}, \bibinfo{person}{Diansheng Guo}, {and} \bibinfo{person}{Depeng Jin}.} \bibinfo{year}{2021}\natexlab{}.
\newblock \showarticletitle{3dgcn: 3-dimensional dynamic graph convolutional network for citywide crowd flow prediction}.
\newblock \bibinfo{journal}{\emph{ACM Transactions on Knowledge Discovery from Data (TKDD)}} \bibinfo{volume}{15}, \bibinfo{number}{6} (\bibinfo{year}{2021}), \bibinfo{pages}{1--21}.
\newblock


\bibitem[Yao et~al\mbox{.}(2021)]%
        {yao2021survey}
\bibfield{author}{\bibinfo{person}{Liuyi Yao}, \bibinfo{person}{Zhixuan Chu}, \bibinfo{person}{Sheng Li}, \bibinfo{person}{Yaliang Li}, \bibinfo{person}{Jing Gao}, {and} \bibinfo{person}{Aidong Zhang}.} \bibinfo{year}{2021}\natexlab{}.
\newblock \showarticletitle{A survey on causal inference}.
\newblock \bibinfo{journal}{\emph{ACM Transactions on Knowledge Discovery from Data (TKDD)}} \bibinfo{volume}{15}, \bibinfo{number}{5} (\bibinfo{year}{2021}), \bibinfo{pages}{1--46}.
\newblock


\bibitem[Yu et~al\mbox{.}(2017)]%
        {yu2017spatio}
\bibfield{author}{\bibinfo{person}{Bing Yu}, \bibinfo{person}{Haoteng Yin}, {and} \bibinfo{person}{Zhanxing Zhu}.} \bibinfo{year}{2017}\natexlab{}.
\newblock \showarticletitle{Spatio-temporal graph convolutional networks: A deep learning framework for traffic forecasting}.
\newblock \bibinfo{journal}{\emph{arXiv preprint arXiv:1709.04875}} (\bibinfo{year}{2017}).
\newblock


\bibitem[Zhang et~al\mbox{.}(2017)]%
        {zhang2017deep}
\bibfield{author}{\bibinfo{person}{Junbo Zhang}, \bibinfo{person}{Yu Zheng}, {and} \bibinfo{person}{Dekang Qi}.} \bibinfo{year}{2017}\natexlab{}.
\newblock \showarticletitle{Deep spatio-temporal residual networks for citywide crowd flows prediction}. In \bibinfo{booktitle}{\emph{Proceedings of the AAAI conference on artificial intelligence}}, Vol.~\bibinfo{volume}{31}.
\newblock


\bibitem[Zhang et~al\mbox{.}(2023)]%
        {zhang2023promptst}
\bibfield{author}{\bibinfo{person}{Zijian Zhang}, \bibinfo{person}{Xiangyu Zhao}, \bibinfo{person}{Qidong Liu}, \bibinfo{person}{Chunxu Zhang}, \bibinfo{person}{Qian Ma}, \bibinfo{person}{Wanyu Wang}, \bibinfo{person}{Hongwei Zhao}, \bibinfo{person}{Yiqi Wang}, {and} \bibinfo{person}{Zitao Liu}.} \bibinfo{year}{2023}\natexlab{}.
\newblock \showarticletitle{Promptst: Prompt-enhanced spatio-temporal multi-attribute prediction}. In \bibinfo{booktitle}{\emph{Proceedings of the 32nd ACM International Conference on Information and Knowledge Management}}. \bibinfo{pages}{3195--3205}.
\newblock


\bibitem[Zhao et~al\mbox{.}(2016)]%
        {zhao2016urban}
\bibfield{author}{\bibinfo{person}{Kai Zhao}, \bibinfo{person}{Sasu Tarkoma}, \bibinfo{person}{Siyuan Liu}, {and} \bibinfo{person}{Huy Vo}.} \bibinfo{year}{2016}\natexlab{}.
\newblock \showarticletitle{Urban human mobility data mining: An overview}. In \bibinfo{booktitle}{\emph{2016 IEEE International Conference on Big Data (Big Data)}}. IEEE, \bibinfo{pages}{1911--1920}.
\newblock


\bibitem[Zhiwen et~al\mbox{.}(2023)]%
        {zhiwen2023assessing}
\bibfield{author}{\bibinfo{person}{Zhang Zhiwen}, \bibinfo{person}{Hongjun Wang}, \bibinfo{person}{Zipei Fan}, \bibinfo{person}{Ryosuke Shibasaki}, {and} \bibinfo{person}{Xuan Song}.} \bibinfo{year}{2023}\natexlab{}.
\newblock \showarticletitle{Assessing the Continuous Causal Responses of Typhoon-related Weather on Human Mobility: An Empirical Study in Japan}. In \bibinfo{booktitle}{\emph{Proceedings of the 32nd ACM International Conference on Information and Knowledge Management}}. \bibinfo{pages}{3524--3533}.
\newblock


\end{thebibliography}

\appendix
\section{Table for Notations}
In this section, we will provide a table of notation used in our methodology in Table~\ref{notation}.

\begin{table}[]
\centering
\caption{Notation used in methodology}
\begin{tabular}{|l|p{6cm}|}
\hline
 $\textbf{P}_{i}$ &  POI distributions in i-th selected region, we categorized POI into 17 types \\ \hline
 $\textbf{X}^{flow}_{i,t}$& Historical human mobility information of $i$-th region at $t$-th time step within input window  \\ \hline
 $\textbf{X}_{t}$ & Temporal information with a input window \\ \hline
 $A_{t}$& Normalized human mobility network among selected regions \\ \hline
 $\textbf{z}_{i,t}$ & The representations of learned spatio-temporal confounders of $i$-th region at $t$-th time step \\ \hline
 $T_{i,t}$& The encoded treatment varibles of $i$-th region at $t$-th time step, fusing all historical events in input window and predictable ones in the output window \\ \hline
 $w_{i,t}$& The output of re-weighting network the balance the selection bias of $i$-th region at $t$-th time step based on $\textbf{z}$ and $T$ \\ \hline
 $N_{i,t}$ & Human intentions toward public events in $i$-th region at $t$-th time step, all-zero scores will be used in there is no event \\ \hline
\end{tabular}
\label{notation}
\end{table}
\section{Hyper-Parameter Settings}\label{hyper}
In this section, we provide other hyper-parameters for model training that we did not mention in the main content because of space limitation.
\begin{table}[htbp]
\caption{Additional hyper-parameter information}
\centering
\begin{tabular}{|l|l|}
\hline
learning rate decay rate &  0.5\\ \hline
early stop threshold & 9e-6 \\ \hline
patience step & 5 \\ \hline
hidden state dimension &  64 \\ \hline
maximum training epoch & 300 \\ \hline
learning rate scheduler metric & \textit{MSE} \\ \hline
random seeds & [1111, 2222, 3333, 4444, 5555] \\ \hline
batch size &  24 \\ \hline
\end{tabular}
\end{table}
\section{News Articles with LLMs Generated Answers}\label{newscases}
In this section, we provide some cases of news articles and their structured information generated with LLMs. Due to copyright and other policies, we cannot share original articles, but we can provide links to the corresponding news articles (English version, not original Japanese news article).\\
\href{https://english.kyodonews.net/news/2023/08/ebf5d8832d49-powerful-typhoon-approaches-japans-okinawa-leaves-1-dead.html}{\color{blue}Typhoon Khanun}:\\
\textbf{Answer 1}: I've identified the most influential events in the news content: Typhoon No. 6 is expected to approach Okinawa on August 3, (predicted closest approach).\\
\textbf{Answer 2}: 2023-08-03 06:00:00\\
\textbf{Answer 3}: Here are my evaluations:\\
\indent Where: People in Okinawa and Amami regions may need to move to safer areas or evacuation centers, away from coastal areas prone to storm surges and flooding.\\
\indent Who: Residents, commuters, and tourists in Okinawa and Amami regions, especially those living in low-lying areas or near the coast, will be affected by the typhoon.\\
\indent When: The event is ongoing, with the typhoon expected to approach Okinawa on August 3. (Note: The news release time was August 2, 07:17:28)\\
\indent How: If necessary, people should move to safer areas or evacuation centers using safe routes and modes of transportation, such as official evacuation buses or private vehicles, while following instructions from local authorities and staying informed about weather updates.\\
\textbf{Answer 4}: No. \\
\textbf{Answer 5}: [80, 20, 0, 0, 40, 70, 80, 60, 30, 50]\\
\href{https://english.kyodonews.net/news/2023/07/1b5901f911f7-major-tokyo-fireworks-festival-resumes-with-a-bang-after-covid-hiatus.html}{\color{blue}Sumidagawa Firework}:\\
\textbf{Answer 1}: The most influential event identified is the resumption of the "Sumida River Fireworks Festival" after a 4-year hiatus due to the COVID-19 pandemic. This event will take place on July 29th and is expected to attract approximately 950,000 people.\\
\textbf{Answer 2}: 2023-07-29 19:07:04\\
\textbf{Answer 3}: Here are my evaluations:\\
\indent Where: If necessary, people may need to move away from the Sumida River area in Tokyo to avoid congestion and large crowds.\\
\indent Who: People who plan to travel or move around in the Sumida River area in Tokyo, particularly those who live or work nearby, tourists, and festival-goers will be affected by the event.\\
\indent When: The event happened on July 29th (the day of the news release).\\
\indent How: If necessary, people should consider using public transportation, rescheduling their travel plans, and planning alternative routes to avoid congested roads and traffic jams.\\
\textbf{Answer 4}: No. \\
\textbf{Answer 5}: [0, 20, 85, 20, 0, 0, 0, 0, 0, 60]

\section{About publicity of source code and data}
We have already open-sourced the code for this research on GitHub, which can be accessed at \href{https://github.com/YangXiaojie1998/CausalMob/}{\color{blue}this github repository}. However, due to data privacy policies, we cannot share human mobility data and the original text-based news articles (though the news articles are publicly available). To ensure that our source code can be implemented smoothly, we synthesized a portion of the data with additional noise, enabling our source code to be run without issues.

\section{Prompts for LLMs}\label{prompt}
In this section, we list all the prompts used to extract structured information and human intentions from news articles. Specifically, we first input the release time of the news (to assist LLMs in inferring the timing of public events), along with the title and content of the news. We then engage the LLMs in a dialogue-based chain-of-thought paradigm by sequentially asking the following questions. This process ultimately generates human intentions, which serve as treatment variables in our causal inference framework.

\begin{promptframed}{Prompt 1: Public Events}
You are an AI assistant that notifies affected users and makes suggestions to change their mobility based on news text.
First, identify the most influential events in the news content, including scheduled and unpredictable events. 
\end{promptframed} 
\noindent
\begin{promptframed}{Prompt 2: Time Information}
Next, estimate the exact time of the most important event mentioned in the news in the following JSON format: “event time”: \textit{“yyyy-mm-dd hh:mm:ss”}. If the exact time is unknown, use the news release time. Provide only the JSON string without any additional text or explanations.
\end{promptframed}
\begin{promptframed}{Prompt 3: 3W1H about Human Mobility}
Based on the news text and our chat, evaluate the 3W1H related to human mobility.\\
- Where: Where should people move if necessary?\\
- Who: What kind of people will be affected?\\
- When: When did the event happen?\\
- How: How should people move if necessary?
\end{promptframed}
\begin{promptframed}{Prompt 4: Predictability}
Is the content in the news more like an unpredictable event, such as an earthquake? Your answer can only be “Yes” or “No”.
\end{promptframed}
\begin{promptframed}{Prompt 5: Human Intentions}
Background: Most news related to economic, politics, culture, and history issues usually have no effect on human mobility because they do not relate to people’s daily life.\\
- Slight disasters, such as light earthquakes and tsunamis, or some local events (like political events) may also have no effect on human mobility in Japan.\\
- Only events that happened close to the release time (within several hours) may have an influence on human mobility.\\
Task: Score the news text based on the following aspects (0-100, where a higher number means higher agreement):\\
Q1. To what extent do the events described in the news make people leave the area because they are dangerous?\\
Q2. To what extent do the events described in the news make people stay in the area because it is better not to move?\\
Q3. To what extent do the events described in the news make people visit the area because they are interesting events?\\
Q4. To what extent do the events described in the news make people keep their daily routine as these events are not important to daily life?\\
Q5. To what extent do the events described in the news lead to interruption of economic activities, such as business closures or work stoppages?\\
Q6. To what extent do the events described in the news affect transportation conditions, such as traffic congestion or road closures?\\
Q7. To what extent do the events described in the news impact public health and safety, leading to decisions to leave or avoid certain areas?\\
Q8. To what extent do the events described in the news involve government or official instructions that influence people’s movements?\\
Q9. To what extent do the events described in the news affect the availability of public services, such as school closures or interruptions in medical services?\\
Q10. To what extent do the events described in the news last a long time (like one day)?\\
Expected response: A list of 10 numbers between 0 and 100.
\end{promptframed}

\end{document}